\documentclass[10pt,twocolumn,letterpaper]{article}

\usepackage[]{cvpr}

\usepackage[utf8]{inputenc}
\usepackage{graphicx}
\usepackage{amsmath,amssymb,amsfonts, amsthm}
\usepackage{times}
\usepackage{epsfig}
\usepackage{array}
\usepackage{caption}
\usepackage{subcaption}
\usepackage{mdframed}
\usepackage{booktabs}
\usepackage{multirow}
\usepackage[pagebackref,breaklinks,colorlinks]{hyperref}
\usepackage{comment}

\usepackage[capitalize]{cleveref}
\crefname{section}{Sec.}{Secs.}
\Crefname{section}{Section}{Sections}
\Crefname{table}{Table}{Tables}
\crefname{table}{Tab.}{Tabs.}

\def\cvprPaperID{} 
\def\confName{}
\def\confYear{2022}

\newcommand{\norm}[1]{\left\lVert #1 \right\rVert}
\newtheorem{theorem}{Theorem}
\newtheorem{lemma}{Lemma}
\newtheorem{corollary}{Corollary}
\newtheorem{assumption}{Assumption}

\newenvironment{customthm}[1]
  {\renewcommand\thetheorem{#1}\theorem}
  {\endtheorem}
\newenvironment{customcoll}[1]
  {\renewcommand\thecorollary{#1}\corollary}
  {\endcorollary}

\makeatletter
\newcommand{\subalign}[1]{%
  \vcenter{%
    \Let@ \restore@math@cr \default@tag
    \baselineskip\fontdimen10 \scriptfont\tw@
    \advance\baselineskip\fontdimen12 \scriptfont\tw@
    \lineskip\thr@@\fontdimen8 \scriptfont\thr@@
    \lineskiplimit\lineskip
    \ialign{\hfil$\m@th\scriptstyle##$&$\m@th\scriptstyle{}##$\hfil\crcr
      #1\crcr
    }%
  }%
}
\makeatother

\usepackage{xcolor}

\DeclareMathOperator*{\argmin}{arg\,min}

\title{Conditional Image Generation with Score-Based Diffusion Models}

\author{Georgios Batzolis\\
DAMTP, \ University of Cambridge\\
{\tt\small gb511@cam.ac.uk}
\and
Jan Stanczuk\\
DAMTP, \ University of Cambridge\\
{\tt\small js2164@cam.ac.uk}
\and
Carola-Bibiane Schönlieb\\
DAMTP, \ University of Cambridge\\
{\tt\small cbs31@cam.ac.uk}
\and
Christian Etmann\\
Deep Render\\
{\tt\small christian.etmann@deeprender.ai}
}

\date{October 2021}

\makeatletter
\DeclareRobustCommand
  \myvdots{\vbox{\baselineskip4\p@ \lineskiplimit\z@
    \hbox{.}\hbox{.}\hbox{.}}}
\makeatother

\DeclareRobustCommand
  \Compactcdots{\mathinner{\cdotp\mkern-2mu\cdotp\mkern-2mu\cdotp}}
  
\begin{document}

\maketitle

\begin{abstract}
    Score-based diffusion models have emerged as one of the most promising frameworks for deep generative modelling. In this work we conduct a systematic comparison and theoretical analysis of different approaches to learning conditional probability distributions with score-based diffusion models. In particular, we prove results which provide a theoretical justification for one of the most successful estimators of the conditional score. Moreover, we introduce a multi-speed diffusion framework, which leads to a new estimator for the conditional score, performing on par with previous state-of-the-art approaches. Our theoretical and experimental findings are accompanied by an open source library \texttt{MSDiff} which allows for application and further research of multi-speed diffusion models.
\end{abstract}

\begin{figure}[h]
    \begin{center}
    \begin{tabular}{ccc}
        \scriptsize Original image $x$ & \scriptsize  Observation $y$ &  \scriptsize  Sample from $p_\theta(x|y)$  \\

        \includegraphics[width=.13\textwidth]{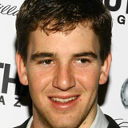} &   
        \includegraphics[width=.13\textwidth]{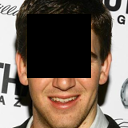} &
        \includegraphics[width=.13\textwidth]{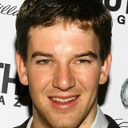}  \\

        \includegraphics[width=.13\textwidth]{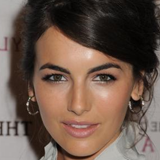} &   
        \includegraphics[width=.13\textwidth]{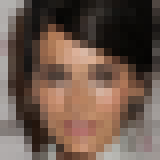} &
        \includegraphics[width=.13\textwidth]{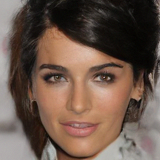}  \\

        \includegraphics[width=.13\textwidth]{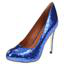} &   
        \includegraphics[width=.13\textwidth]{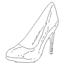} &
        \includegraphics[width=.13\textwidth]{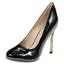}  \\
    \end{tabular}
    \end{center}
    \caption{Results from our conditional multi-speed diffusive estimator.}
    \label{fig: teaser}
\end{figure}

\section{Introduction}
The goal of generative modelling is to learn a  probability distribution from a finite set of samples. This classical problem in statistics has been studied for many decades, but until recently efficient learning of high-dimensional distributions remained impossible in practice. For images, the strong inductive biases of convolutional neural networks have recently enabled the modelling of such distributions, giving rise to the field of deep generative modelling.

Deep generative modelling became one of the central areas of deep learning with many successful applications.
In recent years much progress has been made in unconditional and conditional image generation.
The most prominent approaches are auto-regressive models \cite{bengio2005autoregressive}, variational auto-encoders (VAEs) \cite{kingma2014autoencoding},  normalizing flows \cite{papamakarios2021normalizing} and generative adversarial networks (GANs) \cite{goodfellow2014generative}.

Despite their success, each of the above methods suffers from important limitations. Auto-regressive models allow for likelihood estimation and high-fidelity image generation, but require a lot of computational resources and suffer from poor time complexity in high resolutions. VAEs and normalizing flows are less computationally expensive and allow for likelihood estimation, but tend to produce samples of lower visual quality. Moreover, normalizing flows put restrictions on the possible model architectures (requiring invertibility of the network and a Jacobian log-determinant that is computationally tractable), thus limiting their expressivity. While GANs produce state-of-the art quality samples, they don't allow for likelihood estimation and are notoriously hard to train due to training instabilities and mode collapse. 

Recently, score-based \cite{hyvarinen2005score_original} and diffusion-based  \cite{sohldickstein2015diffusion_original} generative models have been revived and improved in \cite{song2020generative_score} and \cite{ho2020denoising}.  
The connection between the two frameworks in discrete-time formulation has been discovered in \cite{vincent2011connection}. 
Recently in \cite{song2021sde}, both frameworks have been unified into a single continuous-time approach based on stochastic differential equations \cite{song2021sde} and called score-based diffusion models. 
These approaches have recently received a lot of attention, achieving state-of-the-art performance in likelihood estimation \cite{song2021sde} and unconditional image generation \cite{dhariwal2021diffusion_beats_gans}, surpassing even the celebrated success of GANs.

In addition to achieving state-of-the art performance in both image generation and likelihood estimation, score-based diffusion models don't suffer from training instabilities or mode collapse \cite{dhariwal2021diffusion_beats_gans, song2021sde}. Moreover, their time complexity in high resolutions is much better than that of auto-regressive models \cite{dhariwal2021diffusion_beats_gans}. This makes score-based diffusion  models very attractive for deep generative modelling.

In this work, we examine how score-based diffusion  models can be applied to conditional image generation. We conduct a review and classification of existing approaches and perform a systematic comparison to find the best way of estimating the conditional score. We provide a proof of validity for the \textit{conditional denoising estimator} (which has been used in \cite{saharia2021sr3,tashiro2021csdi} without justification), and we thereby provide a firm theoretical foundation for using it in future research.

Moreover, we extend the original framework to support \textit{multi-speed diffusion}, where different parts of the input tensor diffuse according to different speeds. This allows us to introduce a novel estimator of the conditional score and opens an avenue for further research.

\noindent
\textbf{The contributions of this paper are as follows:}
\begin{enumerate}
    \item We review and empirically compare score-based diffusion  approaches to modelling conditional distributions of image data. The models are evaluated on the tasks of super-resolution, inpainting and edge to image translation.
    \item We provide a proof of consistency for the \textit{conditional denoising estimator} - one of the most successful approaches to estimating the conditional score. 
    \item We introduce a multi-speed diffusion framework which leads to \textit{conditional multi-speed diffusive estimator} (CMDE), a novel estimator of conditional score, which unifies previous methods of conditional score estimation.
    \item We provide an open-source library \texttt{MSDiff}, to facilitate further research on conditional and multi-speed diffusion models. \footnote{The code will be released in the near future.}
\end{enumerate}

\section{Notation}

In this work we will use the following notation:
\begin{itemize}
    \item \textbf{Functions of time}
    \begin{gather*}
        f_t := f(t)
    \end{gather*}
    \item \textbf{Indexing vectors} \\
    Let $v = (v_1, ..., v_n) \in \mathbb{R}^n$ and let $ 1 \leq i < j < n$. Then:
    \begin{align*}
        v[:j] &:= (v_1, v_{2}, ..., v_j) \in \mathbb{R}^{j},
    \end{align*}
    cf. Section \ref{sec:CDiffE}.
    \item \textbf{Probability distributions} \\
    We denote the probability distribution of a random variable solely via the name of its density's argument, e.g.
    \begin{gather*}
        p(x_t) := p_{X_t}(x_t),
    \end{gather*}   
    where $x_t$ is a realisation of the random variable $X_t$.
    \item \textbf{Iterated Expectations}
    \begin{align*}
        &\mathbb{E}_{\subalign{z_1 &\sim p(z_1) \\ &{\myvdots} \\z_n &\sim p(z_n)}}[f(z_1,\Compactcdots,z_n)] \\
        :=&\mathbb{E}_{z_1 \sim p(z_1)} \Compactcdots \mathbb{E}_{z_n \sim p(z_n)}[f(z_1,\Compactcdots,z_n)]
    \end{align*}

\end{itemize} 

\section{Methods}
In the following, we will provide details about the framework and estimators discussed in this paper.
\subsection{Background: Score matching through Stochastic Differential Equations}
\subsubsection{Unconditional generation}

In a recent work \cite{song2021sde} score-based  \cite{hyvarinen2005score_original, song2020generative_score} and diffusion-based \cite{sohldickstein2015diffusion_original, ho2020denoising} generative models have been unified into a single continuous-time score-based framework with diffusion driven by stochastic differential equations.  This continuous-time score-based diffusion technique relies on Anderson's Theorem \cite{anderson1982reverse_time_sde}, which states that (under certain assumptions on $\mu : \mathbb{R}^{n_x} \times \mathbb{R} \xrightarrow{} \mathbb{R}^{n_x}$ and $\sigma : \mathbb{R} \xrightarrow{} \mathbb{R}$) a forward diffusion process
\begin{gather}
\label{eq:forward_sde}
 dx = \mu(x,t)dt+\sigma(t)dw   
\end{gather} 
has a reverse diffusion process governed by the following SDE:
\begin{gather}
\label{eq:reverse_sde}
    dx=[\mu(x,t)-\sigma(t)^{2}\nabla_{x}{\ln{p_{X_t}(x)}}]dt + \sigma(t)d\Bar{w},
\end{gather}
where $\Bar{w}$ is a standard Wiener process in reverse time. 

The forward diffusion process transforms the \textit{target distribution} $p(x_0)$ to a \textit{diffused distribution} $p(x_T)$. By appropriately selecting the drift and the diffusion coefficients of the forward SDE, we can make sure that after sufficiently long time $T$, the diffused distribution $p(x_T)$ approximates a simple distribution, such as $\mathcal{N}(0,I)$. We refer to this simple distribution as the \textit{prior distribution}, denoted by $\pi$. 

If we have access to the score of the marginal distribution, $\nabla_{x_t}{\ln{p(x_t)}}$, for all $t$, we can derive the reverse diffusion process and simulate it to map $p_T$ to $p_0$. In practice, we approximate the score of the time-dependent distribution by a neural network $s_{\theta}(x_t,t) \approx \nabla_{x_t}{\ln{p(x_t)}}$ and map the prior distribution $\pi \approx p(x_T)$ to $p_\theta(x) \approx p(x_0)$ by solving the reverse-time SDE from time $T$ to time $0$. One can integrate the reverse SDE using standard numerical SDE solvers such Euler–Maruyama or other discretisation strategies. The authors propose to couple the standard integration step with a fixed number of Langevin MCMC steps to leverage the knowledge of the score of the distribution at each intermediate timestep. The MCMC correction step improves sampling; the combined algorithm is known as a predictor-corrector scheme.  We refer to \cite{song2021sde} for details.

In order to fit a neural network model $s_\theta(x_t,t)$ to approximate the score $\nabla_{x_t}{\ln{p(x_t)}}$, we minimize the weighted Fisher's divergence
\begin{gather}
    \mathcal{L}_{SM}(\theta) := \frac{1}{2} \mathbb{E}_{\subalign{&t \sim U(0,T)\\ &x_t \sim p(x_t)}} [\lambda(t) \norm{\nabla_{x_t}{\ln{p(x_t)}} - s_\theta(x_t,t)}_2^2]
\end{gather}
where $\lambda: [0,T] \xrightarrow{} \mathbb{R}_+$ is a positive weighting function.

The above quantity cannot be optimized directly since we don't have access to the ground truth score $\nabla_{x_t}{\ln{p(x_t)}}$. Therefore in practice, a different objective has to be used \cite{hyvarinen2005score_original, song2020generative_score, song2021sde}. In \cite{song2021sde}, the continuous denoising score-matching objective is chosen, which is equal to $\mathcal{L}_{SM}(\theta)$ up to an additive term, which does not depend on $\theta$ and is defined as 
\begin{gather}
\begin{aligned}
    &\mathcal{L}_{DSM}(\theta) := \\ 
    &\frac{1}{2} \mathbb{E}_{\subalign{&t \sim U(0,T)\\ &x_0 \sim p(x_0) \\ &x_t \sim p(x_t | x_0)}} [\lambda(t) \norm{\nabla_{x_t}{\ln{p(x_t | x_0)}} - s_\theta(x_t,t)}_2^2]
\end{aligned}
\end{gather}
The above expression involves only $\nabla_{x_t}{\ln{p(x_t | x_0)}}$ which can be computed analytically from the transition kernel of the forward diffusion process (provided that $\mu$ and $\sigma$ are sufficiently simple). The expectation can be approximated using Monte Carlo estimation in the following way: First, we sample time points $t_i$ from the uniform distribution on $[0,T]$, then we sample points $\tilde{x}_i$ from the target distribution $p_0$ (available via training set). Next, for each $\tilde{x}$, we sample points $x_i$ from the transition kernel $p(x_t | \tilde{x})$. Finally, we average the expression inside the expectation over all samples obtained in this way.

\subsection{Conditional generation}
The continuous score-matching framework can be extended to conditional generation, as shown in  \cite{song2021sde}. Suppose we are interested in $p(x|y)$, where $x$ is a \textit{target image} and $y$ is a \textit{condition image}. Again, we use the forward diffusion process (Equation \ref{eq:forward_sde}) to obtain a family of diffused distributions $p(x_t | y)$ and apply Anderson's Theorem to derive the \textit{conditional reverse-time SDE}
\begin{equation}
    \label{eq:conditional_reverse_sde}
    dx = [\mu(x,t) - \sigma(t)^2 \nabla_{x} \ln p_{X_t}(x | y)]dt + \sigma(t)d\tilde{w}.
\end{equation}
Now we need to learn the score $\nabla_{x_t} \ln p(x_t|y)$ in order to be able to sample from $p(x | y)$ using reverse-time diffusion.\\

In this work, we discuss the following approaches to estimating the conditional score $\nabla_{x_t} \ln p(x_t|y)$:
\begin{enumerate}
    \item Conditional denoising estimators
    \item Conditional diffusive estimators
    \item Multi-speed conditional diffusive estimators (our method)
\end{enumerate}
We discuss each of them in a separate section.\\

In \cite{song2021sde} an additional approach to conditional score estimation was suggested:
This method proposes learning $\nabla_{x_t} \ln p(x_t)$ with an unconditional score model, and  learning $p(y | x_t)$ with an auxiliary model. Then, one can use 
    \begin{equation*}
        \nabla_{x_t}\ln p(x_t |y) = \nabla_{x_t} \ln p(x_t) + \nabla_{x_t} \ln p(y | x_t)
    \end{equation*}
to obtain $\nabla_{x_t} \ln p(x_t | y)$. Unlike other approaches, this requires training a separate model for $p(y|x_t)$. Appropriate choices of such models for tasks discussed in this paper have not been explored yet. Therefore we exclude this approach from our study. 

\subsubsection{Conditional denoising estimator (CDE)}
The conditional denoising estimator (CDE) is a way of estimating $p(x_t|y)$ using the denoising score matching approach \cite{vincent2011connection, song2020generative_score}. In order to approximate $p(x_t|y)$, the conditional denoising estimator minimizes
\begin{gather}
\begin{aligned}
        \label{CDN}
        &\frac{1}{2} \mathbb{E}_{\subalign{&t \sim U(0,T)\\ &x_0, y \sim p(x_0, y) \\ &x_t \sim p(x_t | x_0)}} 
        [\lambda(t) \norm{\nabla_{x_t} \ln{p(x_t | x_0)} - s_\theta(x_t, y, t)}_2^2]
\end{aligned}
\end{gather}
This estimator has been shown to be successful in previous works \cite{saharia2021sr3,tashiro2021csdi}, also confirmed in our experimental findings (cf. Section \ref{sec:experiments}). 

Despite the practical success, this estimator has previously been used without a theoretical justification of why training the above objective yields the desired conditional distribution. Since $p(x_t|y)$ does not appear in the training objective, it is not obvious that the minimizer approximates the correct quantity. 

By extending the arguments of \cite{vincent2011connection}, we provide a formal proof that the minimizer of the above loss does indeed approximate the correct conditional score $p(x_t|y)$. This is expressed in the following theorem.

\begin{theorem}
    \label{thm:CDE_consistency}
    The minimizer (in $\theta$) of
    \begin{gather*}
    \begin{aligned}
            \frac{1}{2} \mathbb{E}_{\subalign{&t \sim U(0,T)\\ &x_0, y \sim p(x_0, y) \\ &x_t \sim p(x_t | x_0)}} 
            [\lambda(t) \norm{\nabla_{x_t} \ln{p(x_t | x_0)} - s_\theta(x_t, y, t)}_2^2]
    \end{aligned}
    \end{gather*}    
    is the same as the minimizer of 
    \begin{gather*}
        \frac{1}{2} \mathbb{E}_{\subalign{&t \sim U(0,T)\\ &x_t, y \sim p(x_t, y)}} 
        [\lambda(t) \norm{\nabla_{x_t} \ln{p(x_t | y)} - s_\theta(x_t, y,t)}_2^2]
    \end{gather*}
\end{theorem}
\noindent
The proof for this statement can be found in Appendix \ref{appendix:minimizers}. 
Using the above theorem, the consistency of the estimator can be established.
\begin{corollary}
    Let $\theta^\ast$ be a minimizer of a Monte Carlo approximation of (\ref{CDN}), then (under technical assumptions, cf. Appendix \ref{appendix:consistency}) the conditional denoising estimator $s_{\theta^\ast}(x,y,t)$ is a consistent estimator of the conditional score $\nabla_{x_t} \ln p(x_t | y)$, i.e.
    \begin{gather*}
        s_{\theta^\ast}(x,y,t) \overset{P}{\to} \nabla_{x_t} \ln p(x_t | y)   
    \end{gather*}
    as the number of Monte Carlo samples approaches infinity.
\end{corollary}
\noindent
This follows from the previous theorem and the uniform law of large numbers. Proof in the Appendix \ref{appendix:consistency}.

\subsubsection{Conditional diffusive estimator (CDiffE)}
\label{sec:CDiffE}
Conditional diffusive estimators (CDiffE) have first been suggested in \cite{song2021sde}. The core idea is that instead of learning $p(x_t | y)$ directly, we diffuse both $x$ and $y$ and approximate $p(x_t | y_t)$, using the denoising score matching. Just like learning diffused distribution $\nabla_{x_t} \ln p(x_t)$ improves upon direct estimation of $\nabla_{x} \ln p(x)$ \cite{song2020generative_score, song2021sde}, diffusing both the input $x$ and condition $y$, and then learning $\nabla_{x_t} \ln p(x_t | y_t)$ could make optimization easier and give better results than learning  $\nabla_{x_t} \ln p(x_t | y)$ directly.
    
In order to learn $p(x_t | y_t)$, observe that
\begin{gather*}
    \nabla_{x_t}\ln p(x_t | y_t) = \nabla_{x_t}\ln p(x_t, y_t) = \nabla_{z_t}\ln p(z_t)[:n_x],
\end{gather*}
where $z_t := (x_t, y_t)$ and $n_x$ is the dimensionality of $x$. Therefore we can learn the (unconditional) score of the joint distribution $p(x_t, y_t)$ using the denoising score matching objective just like as in the unconditional case, i.e
\begin{gather}
    \label{CDF}
\begin{aligned}
    &\frac{1}{2} \mathbb{E}_{\subalign{&t \sim U(0,T)\\ &z_0 \sim p_0(z_0) \\ &z_t \sim p(z_t | z_0)}} [\lambda(t) \norm{\nabla_{z_t} \ln{p(z_t |z_0)} - s_\theta(z_t,t)}_2^2].
\end{aligned}
\end{gather}
We can then extract our approximation for the conditional score $\nabla_{x_t} \ln p(x_t|y_t)$ by simply taking the first $n_x$ components of $s_\theta(x_t, y_t,t)$.

The aim is to approximate $\nabla_{x_t} \ln p(x_t | y)$ with $\nabla_{x_t} \ln p(x_t|\hat{y_t})$, where $\hat{y}_t$ is a sample from $p(y_t | y)$. Of course this approximation is imperfect and introduces an error, which we call the \textit{approximation error}. CDiffE aims to achieve smaller optimization error by diffusing the condition $y$ and making the optimization landscape easier, at a cost of making this approximation error.

Now in order to obtain samples from the conditional distribution, we sample a point $x_T \sim \pi$ and integrate
\begin{gather*}
    dx = [\mu(x,t) - \sigma(t)^2 \nabla_{x} \ln p_{X_t|Y_t}(x | \hat{y}_t)]dt + \sigma(t)d\tilde{w}
\end{gather*}
from $T$ to $0$, sampling $\hat{y}_t \sim p(y_t | y)$ at each time step.



\subsubsection{Conditional multi-speed diffusive estimator (CMDE)}
\label{sec:CMDE}
\begin{figure}
\begin{mdframed}
    \begin{center}
       \textbf{Sources of error for different estimators}
    \end{center}
    \textbf{CDE}\\
    Optimization error:
    \begin{gather*}
    s_\theta(x,y,t) \approx \nabla_{x_t}\ln p(x_t|y)   
    \end{gather*}
    \textbf{CDiffE and CMDE}\\
    Optimization error:
    \begin{gather*}
        s_\theta(x,y,t) \approx \nabla_{x_t}\ln p(x_t|y_t)   
        \end{gather*}
    Approximation error:
    \begin{gather*}
        \nabla_{x_t}\ln p(x_t|\hat{y}_t) \approx \nabla_{x_t}\ln p(x_t|y)   
        \end{gather*}
    CDiffE aims to achieve smaller optimization error at a cost of higher approximation error. By controlling the diffusion speed of $y$, CMDE tries to find an optimal balance between optimization error and approximation error.
\end{mdframed}
\caption{Sources of error for different estimators}
\label{fig:box}
\end{figure}
In this section we present a novel estimator for the conditional score $\nabla_{x_t} \ln p(x_t | y)$ which we call the \textit{conditional multi-speed diffusive estimator} (CMDE). 

Our approach is based on two insights. Firstly, there is no reason why $x_t$ and $y_t$ in conditional diffusive estimation need to diffuse at the same rate. Secondly, by decreasing the diffusion rate of $y_t$ while keeping the diffusion speed of $x_t$ the same, we can bring $p(x_t |y_t)$ closer to $p(x_t |y)$, at the possible cost of making the optimization more difficult. This way we can \emph{interpolate} between the conditional denoising estimator and the conditional diffusive estimator and find an optimal balance between optimization error and approximation error (cf. Figure \ref{fig:box}). This can lead to a better performance, as indicated by our experimental findings (cf. Section \ref{sec:experiments}).

In our conditional multi-speed diffusive estimator, $x_t$ and $y_t$ diffuse according to SDEs with the same drift but different diffusion rates,
\begin{gather*}
    dx = \mu(x,t)dt+\sigma^x(t)dw  \\
    dy = \mu(y,t)dt+\sigma^y(t)dw.
\end{gather*}

Then, just like in the case of conditional diffusive estimator, we try to approximate the joint score $\nabla_{x_t, y_t} \ln p(x_t, y_t)$ with a neural network. Since $x_t$ and $y_t$ now diffuse according to different SDEs, we need to take this into account and replace the weighting function $\lambda(t):\mathbb{R} \xrightarrow{} \mathbb{R}_+ $ with a positive definite weighting matrix $\Lambda(t): \mathbb{R} \xrightarrow{} \mathbb{R}^{(n_x + n_y) \times (n_x + n_y)}$. Hence, the new training objective becomes
\begin{gather}
    \label{eq:CMDE}
    \begin{aligned}
        \frac{1}{2} \mathbb{E}_{\subalign{&t \sim U(0,T)\\ &z_0 \sim p_0(z_0) \\ &z_t \sim p(z_t | z_0)}} 
        [
            v^T \Lambda(t) v
        ],
    \end{aligned}
\end{gather}
where $v=\nabla_{z_t} \ln{p(z_t |z_0)} - s_\theta(z_t,t)$, $z_t=(x_t,y_t)$.

In \cite{song2021maximum} authors derive a likelihood weighting function $\lambda^{\text{MLE}}(t)$, which ensures that the objective of the score-based model upper-bounds the negative log-likelihood of the data, thus enabling approximate maximum likelihood training of score-based diffusion models. We generalize this result to the multi-speed diffusion case by providing a likelihood weighting matrix $\Lambda^{\text{MLE}}(t)$ with the same properties.

\begin{theorem}
    \label{thm:weightning}
    Let $\mathcal{L}(\theta)$ be the CMDE training objective (Equation \ref{eq:CMDE}) with the following weighting:
    \begin{gather*}
        \Lambda^{\text{MLE}}_{i,j}(t) =  
        \begin{cases} 
            \sigma^x(t)^2, \text{ if } i=j, \ i \leq n_x \\ 
            \sigma^y(t)^2, \text{ if } i=j, \ n_x < i \leq n_y  \\
            0, \text{ otherwise}
        \end{cases}            
    \end{gather*}
    Then the joint negative log-likelihood is upper bounded (up to a constant in $\theta$) by the training objective of CMDE
    \begin{gather*}
        -\mathbb{E}_{(x,y) \sim p(x,y)}[\ln p_\theta(x,y)] \leq \mathcal{L}(\theta) + C.
    \end{gather*}
\noindent
\end{theorem}



\noindent
The proof can be found in Appendix \ref{appendix:weighting}. 

Moreover we show that the mean squared approximation error of a multi-speed diffusion model is upper bounded and the upper bound goes to zero as the diffusion speed of the condition $\sigma^y(t)$ approaches zero.
\begin{theorem}
    Fix $t$, $x_t$ and $y$. Under mild technical assumptions (cf. Appendix \ref{appendix:mse}) there exists a function  $E: \mathbb{R} \xrightarrow{} \mathbb{R}$ monotonically decreasing to $0$, such that
    \begin{gather*}
        \mathbb{E}_{y_t \sim p(y_t|y)}[
            \norm{ \nabla_{x_t} \ln p(x_t|y_t) - \nabla_{x_t} \ln p(x_t|y)}_2^2
            ] \\
            \leq E(1/\sigma^y(t)).
    \end{gather*}
\end{theorem}
\noindent
The proof can be found in Appendix \ref{appendix:mse}.

Thus we see that the objective of CMDE approaches that of CDE as  $\sigma^y(t) \to 0$, and  CMDE coincides with CDiffE when $ \sigma^y(t) = \sigma^x(t)$ (cf. Figure \ref{fig:box}).

We experimented with different configurations of $\sigma^x(t)$ and $\sigma^y(t)$ and found configurations that lead to  improvements upon CDiffE and  CDE in certain tasks. The experimental results are discussed in detail in Section \ref{sec:experiments}.

\subsubsection{\texttt{MSDiff}: Beyond multi-speed diffusion}
\label{sec:multi-sde}

Based on this work, we provide our open source library \texttt{MSDiff}, which generalizes the original framework of \cite{song2021sde} and allows to diffuse $x_t$ and $y_t$ not only at different diffusion rates, but with two entirely different forward SDEs (i.e. with different diffusion coefficients \emph{and} different drifts):
\begin{gather*}
    dx = \mu^x(x,t)dt+\sigma^x(t)dw  \\
    dy = \mu^y(y,t)dt+\sigma^y(t)dw  
\end{gather*}
Moreover, the likelihood weighting of Theorem \ref{thm:weightning} holds in this more general case, allowing for principled training of multi-sde diffusion models (cf. Appendix \ref{appendix:weighting}).
This flexibility opens room for further research into multi-speed diffusion based training of score-based models, which we intend to examine in a future study. 

\section{Experiments}
\label{sec:experiments}

\begin{figure*}
    \captionsetup[subfigure]{labelformat=empty}
    \begin{subfigure}{.135\textwidth}
        \includegraphics[width=\textwidth]{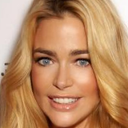}
        \caption{\scriptsize Original image $x$}
    \end{subfigure}
    \begin{subfigure}{.135\textwidth}
        \includegraphics[width=\textwidth]{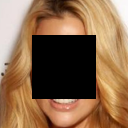}
        \caption{\scriptsize Observation $y := Ax$}
    \end{subfigure}
    \begin{subfigure}{.135\textwidth}
        \includegraphics[width=\textwidth]{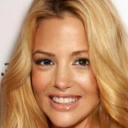}
        \caption{\scriptsize Reconstruction $\hat{x}_1$}
    \end{subfigure} 
    \begin{subfigure}{.135\textwidth}
        \includegraphics[width=\textwidth]{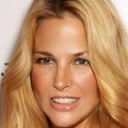}
        \caption{\scriptsize Reconstruction $\hat{x}_2$}
    \end{subfigure}    
    \begin{subfigure}{.135\textwidth}
        \includegraphics[width=\textwidth]{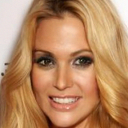}
        \caption{\scriptsize Reconstruction $\hat{x}_3$}
    \end{subfigure}    
    \begin{subfigure}{.135\textwidth}
        \includegraphics[width=\textwidth]{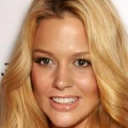}
        \caption{\scriptsize Reconstruction $\hat{x}_4$}
    \end{subfigure}
    \begin{subfigure}{.135\textwidth}
        \includegraphics[width=\textwidth]{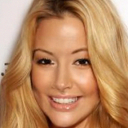}
        \caption{\scriptsize Reconstruction $\hat{x}_5$}
    \end{subfigure} 
    \caption{Diversity of five different CMDE reconstructions for a given compressed image.}
\end{figure*}

In this section we conduct a systematic comparison of different score-based diffusion approaches to modelling conditional distributions of image data. We evaluate these approaches on the tasks of super-resolution, inpainting and edge to image translation.  Moreover, we compare the most successful score-based diffusion approaches for super-resolution with HCFlow  \cite{liang2021hrflow} -- a state-of-the-art method in super-resolution.

\noindent
\textbf{Datasets} In our experiments, we use the CelebA \cite{2015celeba} and Edges2shoes \cite{yu2014sketch2shoe,isola2018pix2pix} datasets. We pre-processed the CelebA dataset as in \cite{liang2021hrflow}.

\noindent
\textbf{Models and hyperparameters} In order to ensure the fair comparison, we separate the evaluation of a particular estimator of conditional score from the evaluation of a particular neural network model. To this end, we train the same neural network architecture for all estimators. The architecture is based on the DDPM model used in \cite{ho2020denoising, song2021sde}. 
We used the variance-exploding SDE \cite{song2021sde} given by: 
$$dx = \sqrt{\frac{d}{dt}\sigma^2(t)}dw, \hspace{1cm}
\sigma(t) = \sigma_{min} \left(\frac{\sigma_{max}}{\sigma_{min}}\right)^t$$
Likelihood weighting was employed for all experiments. For CMDE, the diffusion speed of $y$ was controlled by picking an appropriate $\sigma^y_{\max}$, which we found by trial-and-error. The performance of CMDE could be potentially improved by performing a systematic hyperparameter search for optimal $\sigma^y_{\max}$.
Details on hyperparameters and architectures used in our experiments can be found in Appendix \ref{appendix:hyperparams}.

\noindent 
\textbf{Inverse problems} The tasks of inpainting, super-resolution and edge to image translation are special cases of inverse problems \cite{arridge2019ip, muller2012ip}. In each case, we are given a (possibly random) forward operator $A$ which maps our data $x$ (full image) to an observation $y$ (masked image, compressed image, sketch). The task is to come up with a high-quality reconstruction $\hat{x}$ of the image $x$ based on an observation $y$. The problem of reconstructing $x$ from $y$ is typically ill-posed, since $y$ does not contain all information about $x$. Therefore, an ideal algorithm would produce a reconstruction $\hat{x}$, which looks like a realistic image (i.e. is a likely sample from $p(x)$) and is consistent with the observation $y$ (i.e. $A\hat{x} \approx y$). Notice that if a conditional score model learns the conditional distribution correctly, then our reconstruction $\hat{x}$ is a sample from the posterior distribution $p(x | y)$, which satisfies bespoke requirements. This strategy for solving inverse problems is generally referred to as \emph{posterior sampling}.


\noindent
\textbf{Evaluation: Reconstruction quality} Ill-posedness often means that we should not strive to reconstruct $x$ perfectly. Nonetheless reconstruction error does correlate with the performance of the algorithm and has been one of the most widely-used metrics in the community. To evaluate the reconstruction quality for each task, we measure the Peak signal-to-noise ratio (PSNR) \cite{zhou2004psnr+ssim}, Structural similarity index measure (SSIM) \cite{zhou2004psnr+ssim} and Learned Perceptual Image Patch Similarity (LPIPS) \cite{zhang2018lpips} between the original image $x$ and the reconstruction $\hat{x}$.

\noindent
\textbf{Evaluation: Consistency}
In order to evaluate the consistency of the reconstruction, for each task we calculate the PSNR between $y:=Ax$ and $\hat{y}:=A\hat{x}$. 

\noindent
\textbf{Evaluation: Diversity}
We evaluate diversity of each approach by generating five reconstructions $(\hat{x})_{i=1}^5$ for a given observation ${y}$. Then for each $y$ we calculate the average standard deviation for each pixel among the reconstructions  $(\hat{x})_{i=1}^5$ . Finally, we average this quality over 5000 test observations.

\noindent
\textbf{Evaluation: Distributional distances}
If our algorithm generates realistic reconstructions while preserving diversity, then the distribution of reconstructions $p(\hat{x})$ should be similar to the distribution of original images $p(x)$. Therefore, we measure the Fr\'{e}chet Inception Distance (FID) \cite{heusel2018fid} between unconditional distributions $p(x)$ and $p(\hat{x})$ based on 5000 samples. Moreover, we calculate the FID score between the joint distributions $p(\hat{x}, y)$ and $p(x,y)$, which allows us to simultaneously check the realism of the reconstructions and the consistency with the observation. 
We use abbreviation UFID to refer to FID between between unconditional distributions and JFID to refer to FID between joints.  In our judgement, FID and especially the JFID is the most principled of the used metrics, since it measures how far $p_\theta(x | y)$ is from $p(x|y)$.

\begin{table*}
    \begin{center}
    \caption{Results of conditional generation tasks.}
    \label{tbl:results}
    \begin{tabular}{cccccccc}
    \toprule
    &Estimator & PSNR/SSIM $\uparrow$  & LPIPS $\downarrow$ & UFID/JFID $\downarrow$ & Consistency $\uparrow$ & Diversity $\uparrow$ \\
    \midrule
    \multirow{3}{*}{Inpainting} 
    &CDE & \textbf{25.12}/\textbf{0.870}  & \textbf{0.042} & 13.07/18.06 & \textbf{28.54} & 4.79  \\
    &CDiffE & 23.07/0.844   & 0.057 & 13.28/19.25 &  26.61 & \textbf{6.52}   \\
    &CMDE ($\sigma^y_{max} = 1$) & 24.92/0.864  & 0.044 & \textbf{12.07/17.07} & 28.32 & 4.98  \\
    \midrule
    \multirow{4}{*}{Super-resolution} 
    &CDE & 23.80/0.650  & 0.114 & 10.36/15.77 & 54.18 & \textbf{8.51}  \\
    &CDiffE & 23.83/0.656  & 0.139 & 14.29/20.20 & 51.90 & 7.41  \\
    &CMDE ($\sigma^y_{max} = 0.5$) & 23.91/0.654  & 0.109 & \textbf{10.28/15.68} & 53.03 & 8.33  \\
    &HCFLOW & \textbf{24.95/0.702} & \textbf{0.107} & 14.13/19.55 & \textbf{55.31} & 6.26 \\
    \midrule
    \multirow{3}{*}{Edge to image} 
    &CDE & \textbf{18.35/0.699}  & \textbf{0.156} & \textbf{11.87/21.31} & \textbf{10.45} & 14.40 \\
    &CDiffE & 10.00/0.365   & 0.350 & 33.41/55.22  & 7.78 & \textbf{43.45} \\
    &CMDE ($\sigma^y_{max} = 1$) & 18.16/0.692  & 0.158 & 12.62/22.09 & 10.38 & 15.20  \\
    \bottomrule
    \end{tabular}
    \end{center}
\end{table*}

\subsection{Inpainting}

\begin{figure*}
    \begin{center}
        \begingroup
        \setlength{\tabcolsep}{2pt}
    \begin{tabular}{ccccc}
        Original image $x$ & Observation $y$ & CDE & CDiffE & CMDE (Ours) \\

        \includegraphics[width=.15\textwidth]{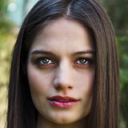} &   
        \includegraphics[width=.15\textwidth]{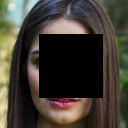} &
        \includegraphics[width=.15\textwidth]{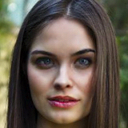} & 
        \includegraphics[width=.15\textwidth]{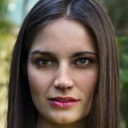} &
        \includegraphics[width=.15\textwidth]{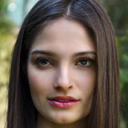} \\

        \includegraphics[width=.15\textwidth]{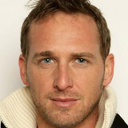} &   
        \includegraphics[width=.15\textwidth]{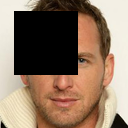} &
        \includegraphics[width=.15\textwidth]{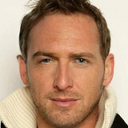} & 
        \includegraphics[width=.15\textwidth]{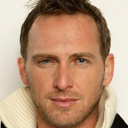} &
        \includegraphics[width=.15\textwidth]{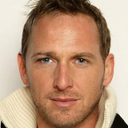} \\
 
        \includegraphics[width=.15\textwidth]{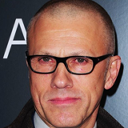} &   
        \includegraphics[width=.15\textwidth]{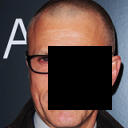} &
        \includegraphics[width=.15\textwidth]{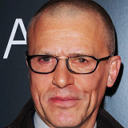} & 
        \includegraphics[width=.15\textwidth]{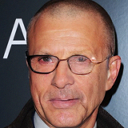} &
        \includegraphics[width=.15\textwidth]{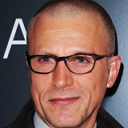} \\
    \end{tabular}
    \endgroup
    \end{center}
    \caption{Inpainting results.}
    \label{fig:inpainting}
\end{figure*}

We perform the inpainting experiment using CelebA dataset. In inpainting, the forward operator $A$ is an application of a given binary mask to an image $x$.  In our case, we made the task more difficult by using randomly placed (square) masks. Then the conditional score model is used to obtain a reconstruction $\hat{x}$ from the masked image $y$. We select the position of the mask uniformly at random and cover $25\%$ of the image. The quantitative results are summarised in Table \ref{tbl:results} and samples are presented in Figure \ref{fig:inpainting}. We observe that CDE and CMDE significantly outperform CDiffE in all metrics, with CDE having a small advantage over CMDE in terms of reconstruction error and consistency. On the other hand, CMDE achieves the best FID scores.


\subsection{Super-resolution}
\begin{figure*}
    \begin{center}
    \begingroup
    \setlength{\tabcolsep}{2pt}

    \begin{tabular}{cccccc}
        Original image $x$ & Observation $ y$ & HCFlow & CDE & CDiffE & CMDE (Ours) \\

        \includegraphics[width=.15\textwidth]{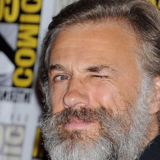} &   
        \includegraphics[width=.15\textwidth]{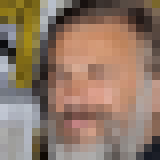} &
        \includegraphics[width=.15\textwidth]{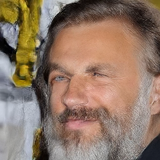} &
        \includegraphics[width=.15\textwidth]{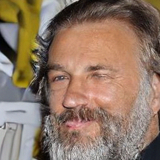} & 
        \includegraphics[width=.15\textwidth]{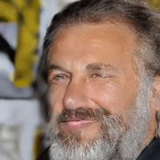} &
        \includegraphics[width=.15\textwidth]{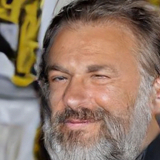} \\

        \includegraphics[width=.15\textwidth]{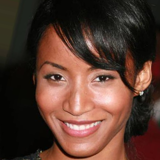} &   
        \includegraphics[width=.15\textwidth]{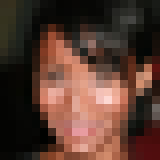} &
        \includegraphics[width=.15\textwidth]{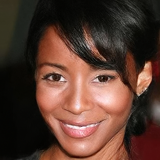} &
        \includegraphics[width=.15\textwidth]{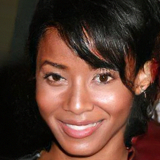} & 
        \includegraphics[width=.15\textwidth]{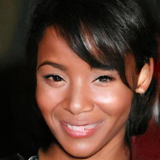} &
        \includegraphics[width=.15\textwidth]{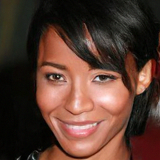} \\

        \includegraphics[width=.15\textwidth]{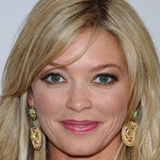} &   
        \includegraphics[width=.15\textwidth]{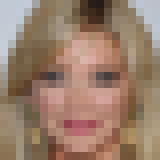} &
        \includegraphics[width=.15\textwidth]{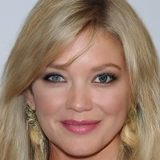} &
        \includegraphics[width=.15\textwidth]{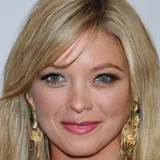} & 
        \includegraphics[width=.15\textwidth]{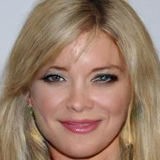} &
        \includegraphics[width=.15\textwidth]{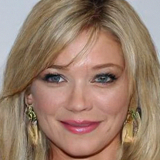} \\
    \end{tabular}
    \endgroup
    \end{center}
    \caption{Super-resolution results.}
    \label{fig:super-resolution}
\end{figure*}

We perform 8x super-resolution using the CelebA dataset. A high resolution 160x160 pixel image $x$ is compressed to a low resolution 20x20 pixels image $y$. Here we use bicubic downscaling \cite{keyes1981bicubic} as the forward operator  $A$. Then using a score model we obtain a 160x160 pixel reconstruction image $\hat{x}$. The quantitative results are summarised in Table \ref{tbl:results} and samples are presented in Figure \ref{fig:super-resolution}. We find that CMDE and CDE perform similarly, while significantly outperforming CDiffE. CMDE achieves the smallest reconstruction error and captures the distribution most accurately according to FID scores.

\subsection{Edge to image translation}
\begin{figure}
\renewcommand{\arraystretch}{1.25}
    \begin{tabular}{lccc}
        \begin{tabular}{@{}l@{}}
            Original image $x$
            \\[25pt]
        \end{tabular}
         & \includegraphics[width=.08\textwidth]{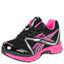} & \includegraphics[width=.08\textwidth]{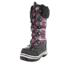} & \includegraphics[width=.08\textwidth]{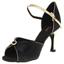} \\
        
        \begin{tabular}{@{}l@{}}
            Observation \\ $ y := Ax$
            \\[25pt]
        \end{tabular}
         & \includegraphics[width=.08\textwidth]{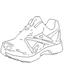} & \includegraphics[width=.08\textwidth]{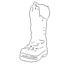} & \includegraphics[width=.08\textwidth]{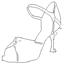} \\
        
         \begin{tabular}{@{}c@{}}
            CDE
            \\[25pt]
        \end{tabular} & \includegraphics[width=.08\textwidth]{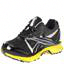}  & \includegraphics[width=.08\textwidth]{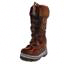}  & \includegraphics[width=.08\textwidth]{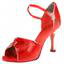} \\

        \begin{tabular}{@{}c@{}}
            CDiffE
            \\[25pt]
        \end{tabular} & \includegraphics[width=.08\textwidth]{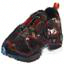} & \includegraphics[width=.08\textwidth]{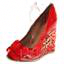} & \includegraphics[width=.08\textwidth]{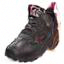}\\

        \begin{tabular}{@{}c@{}}
            CMDE (Ours)
            \\[25pt]
        \end{tabular}  &  \includegraphics[width=.08\textwidth]{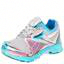} &  \includegraphics[width=.08\textwidth]{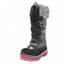} &  \includegraphics[width=.08\textwidth]{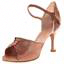} 
    \end{tabular}
    \caption{Edge to image translation results.}
    \label{fig:edges-to-shoes}
\end{figure}

We perform an edge to image translation task on the Edges2shoes dataset. The forward operator $A$ is given by a neural network edge detector \cite{xie2015edges}, which takes an original photo of a shoe $x$ and transforms it into a sketch $y$. Then a conditional score model is used to create an artificial photo of a shoe $\hat{x}$ matching the sketch. The quantitative results are summarised in Table \ref{tbl:results} and samples are presented in Figure \ref{fig:edges-to-shoes}. Unlike in inpainting and super-resolution where CDiffE achieved reasonable performance, in edge to image translation, it fails to create samples consistent with the condition (which leads to inflated diversity scores). CDE and CMDE are comparable, but CDE performed slightly better across all metrics. However, the performance of CMDE could be potentially improved by  tuning the diffusion speed $\sigma^y(t)$.

\section{Comparison with state-of-the-art}
We compare score-based diffusion approaches with HCFlow \cite{liang2021hrflow} -- a state-of-the-art method in super-resolution. To ensure a fair comparison, we used the data pre-processing and hyperparameters for HCFlow exactly as in the original paper \cite{liang2021hrflow}. We find that although HCFlow performs marginally better in terms of the reconstruction error, CDE and CMDE obtain a significantly better FID and diversity scores indicating better distribution coverage. 
We recall that a perfect reconstruction on a per-image basis is generally not desirable due to ill-posedness of the inverse problem and therefore in our view FID is the most principled of the used metrics.
The FID scores suggest that CMDE was the most successful approach to approximating the posterior distribution.

\section{Conclusions and future work}

In this article, we conducted a systematic comparison of score-based diffusion models in conditional image generation tasks and provided an in-depth theoretical analysis of the estimators of conditional score. In particular, we proved the consistency of the conditional denoising estimator, thus providing a firm theoretical justification for using it in future research. 

Moreover, we introduced a multi-speed diffusion framework, which led to CMDE, a novel estimator for conditional score which interpolates between conditional denoising estimator (CDE) and conditional diffusive estimator (CDiffE) by controlling the diffusion speed of the condition.

Our study showed that CMDE and CDE perform on par, while significantly outperforming CDiffE. This is particularly apparent in edge to image translation, where CDiffE fails to produce samples consistent with the condition image. Furthermore, CMDE outperformed CDE in terms of  FID scores in inpainting and super-resolution tasks, which indicates that diffusing the condition at the appropriate speed can have beneficial effect on the optimization landscape, and yield better approximation of the posterior distribution.

We found that score-based diffusion models perform on par with prior state-of-the-art methods in super-resolution task and achieve better posterior approximation according to FID score.


\section{Acknowledgements} 
GB acknowledges the support from GSK and the Cantab Capital Institute for the Mathematics of Information. 
JS acknowledges the support from Aviva and the Cantab Capital Institute for the Mathematics of Information. 
CBS acknowledges support from the Philip Leverhulme Prize, the Royal Society Wolfson Fellowship, the EPSRC advanced career fellowship EP/V029428/1, EPSRC grants EP/S026045/1 and EP/T003553/1, EP/N014588/1, EP/T017961/1, the Wellcome Innovator Award RG98755, the Leverhulme Trust project Unveiling the invisible, the European Union Horizon 2020 research and innovation programme under the Marie Skodowska-Curie grant agreement No. 777826 NoMADS, the Cantab Capital Institute for the Mathematics of Information and the Alan Turing Institute. CE acknowledges support from
the Wellcome Innovator Award RG98755 for part of the work that was done at Cambridge.

{\small
\bibliographystyle{ieee_fullname}
\bibliography{bibliography}

\begin{thebibliography}{10}\itemsep=-1pt

\bibitem{anderson1982reverse_time_sde}
Brian~D.O. Anderson.
\newblock Reverse-time diffusion equation models.
\newblock {\em Stochastic Processes and their Applications}, 12(3):313--326,
  1982.

\bibitem{arridge2019ip}
Simon Arridge, Peter Maass, Ozan Öktem, and Carola-Bibiane Schönlieb.
\newblock Solving inverse problems using data-driven models.
\newblock {\em Acta Numerica}, 28:1–174, 2019.

\bibitem{bengio2005autoregressive}
Yoshua Bengio, R\'{e}jean Ducharme, Pascal Vincent, and Christian Janvin.
\newblock A neural probabilistic language model.
\newblock {\em J. Mach. Learn. Res.}, 3(null):1137–1155, Mar. 2003.

\bibitem{dhariwal2021diffusion_beats_gans}
Prafulla Dhariwal and Alex Nichol.
\newblock Diffusion models beat gans on image synthesis, 2021.

\bibitem{goodfellow2014generative}
Ian~J. Goodfellow, Jean Pouget-Abadie, Mehdi Mirza, Bing Xu, David
  Warde-Farley, Sherjil Ozair, Aaron Courville, and Yoshua Bengio.
\newblock Generative adversarial networks, 2014.

\bibitem{heusel2018fid}
Martin Heusel, Hubert Ramsauer, Thomas Unterthiner, Bernhard Nessler, and Sepp
  Hochreiter.
\newblock Gans trained by a two time-scale update rule converge to a local nash
  equilibrium, 2018.

\bibitem{ho2020denoising}
Jonathan Ho, Ajay Jain, and Pieter Abbeel.
\newblock Denoising diffusion probabilistic models, 2020.

\bibitem{hyvarinen2005score_original}
Aapo Hyv{{\"a}}rinen.
\newblock Estimation of non-normalized statistical models by score matching.
\newblock {\em Journal of Machine Learning Research}, 6(24):695--709, 2005.

\bibitem{isola2018pix2pix}
Phillip Isola, Jun-Yan Zhu, Tinghui Zhou, and Alexei~A. Efros.
\newblock Image-to-image translation with conditional adversarial networks,
  2018.

\bibitem{keyes1981bicubic}
R. Keys.
\newblock Cubic convolution interpolation for digital image processing.
\newblock {\em IEEE Transactions on Acoustics, Speech, and Signal Processing},
  29(6):1153--1160, 1981.

\bibitem{kingma2014autoencoding}
Diederik~P Kingma and Max Welling.
\newblock Auto-encoding variational bayes, 2014.

\bibitem{liang2021hrflow}
Jingyun Liang, Andreas Lugmayr, Kai Zhang, Martin Danelljan, Luc~Van Gool, and
  Radu Timofte.
\newblock Hierarchical conditional flow: A unified framework for image
  super-resolution and image rescaling, 2021.

\bibitem{2015celeba}
Ziwei Liu, Ping Luo, Xiaogang Wang, and Xiaoou Tang.
\newblock Deep learning face attributes in the wild.
\newblock In {\em Proceedings of International Conference on Computer Vision
  (ICCV)}, December 2015.

\bibitem{leonard2013properties}
Christian Léonard.
\newblock Some properties of path measures, 2013.

\bibitem{muller2012ip}
Jennifer~L. Mueller and Samuli Siltanen.
\newblock Linear and nonlinear inverse problems with practical applications.
\newblock In {\em Computational science and engineering}, 2012.

\bibitem{whitney1994estimation}
Whitney~K. Newey and Daniel McFadden.
\newblock Chapter 36 large sample estimation and hypothesis testing.
\newblock volume~4 of {\em Handbook of Econometrics}, pages 2111--2245.
  Elsevier, 1994.

\bibitem{oksendal2003sde}
Bernt Oksendal.
\newblock {\em Stochastic Differential Equations (5th Ed.): An Introduction
  with Applications}.
\newblock Springer-Verlag, Heidelberg, 2003.

\bibitem{papamakarios2021normalizing}
George Papamakarios, Eric Nalisnick, Danilo~Jimenez Rezende, Shakir Mohamed,
  and Balaji Lakshminarayanan.
\newblock Normalizing flows for probabilistic modeling and inference, 2021.

\bibitem{saharia2021sr3}
Chitwan Saharia, Jonathan Ho, William Chan, Tim Salimans, David~J. Fleet, and
  Mohammad Norouzi.
\newblock Image super-resolution via iterative refinement, 2021.

\bibitem{sohldickstein2015diffusion_original}
Jascha Sohl-Dickstein, Eric~A. Weiss, Niru Maheswaranathan, and Surya Ganguli.
\newblock Deep unsupervised learning using nonequilibrium thermodynamics, 2015.

\bibitem{song2021maximum}
Yang Song, Conor Durkan, Iain Murray, and Stefano Ermon.
\newblock Maximum likelihood training of score-based diffusion models, 2021.

\bibitem{song2020generative_score}
Yang Song and Stefano Ermon.
\newblock Generative modeling by estimating gradients of the data distribution,
  2020.

\bibitem{song2021sde}
Yang Song, Jascha Sohl-Dickstein, Diederik~P. Kingma, Abhishek Kumar, Stefano
  Ermon, and Ben Poole.
\newblock Score-based generative modeling through stochastic differential
  equations, 2021.

\bibitem{tashiro2021csdi}
Yusuke Tashiro, Jiaming Song, Yang Song, and Stefano Ermon.
\newblock Csdi: Conditional score-based diffusion models for probabilistic time
  series imputation, 2021.

\bibitem{vincent2011connection}
Pascal Vincent.
\newblock A connection between score matching and denoising autoencoders.
\newblock {\em Neural Computation}, 23(7):1661--1674, 2011.

\bibitem{zhou2004psnr+ssim}
Zhou Wang, A.C. Bovik, H.R. Sheikh, and E.P. Simoncelli.
\newblock Image quality assessment: from error visibility to structural
  similarity.
\newblock {\em IEEE Transactions on Image Processing}, 13(4):600--612, 2004.

\bibitem{xie2015edges}
Saining Xie and Zhuowen Tu.
\newblock Holistically-nested edge detection, 2015.

\bibitem{yu2014sketch2shoe}
A. Yu and K. Grauman.
\newblock Fine-grained visual comparisons with local learning.
\newblock In {\em Computer Vision and Pattern Recognition (CVPR)}, Jun 2014.

\bibitem{zhang2018lpips}
Richard Zhang, Phillip Isola, Alexei~A Efros, Eli Shechtman, and Oliver Wang.
\newblock The unreasonable effectiveness of deep features as a perceptual
  metric.
\newblock In {\em CVPR}, 2018.

\end{thebibliography}
}

\newpage
\appendix
\section{Variance schedule}
\label{appendix:VR}
\begin{figure*}
    \captionsetup[subfigure]{labelformat=empty}
    \begin{subfigure}{.5\textwidth}
        \includegraphics[width=\textwidth]{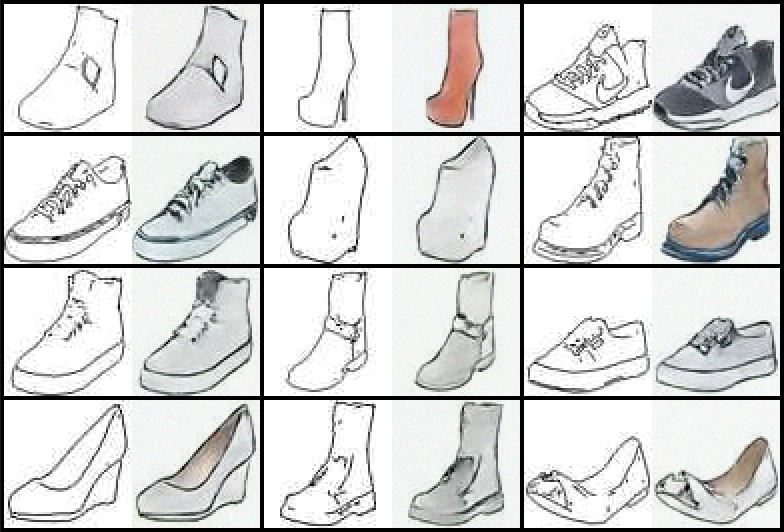}
        \caption{\scriptsize CMDE}
    \end{subfigure}
    \begin{subfigure}{.5\textwidth}
        \includegraphics[width=\textwidth]{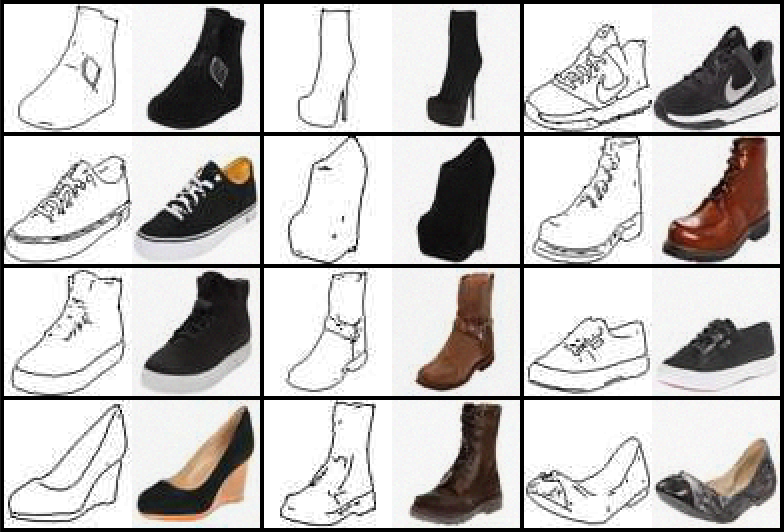}
        \caption{\scriptsize VS-CMDE}
    \end{subfigure}
    \caption{Discrete training. Left: CMDE ($\sigma_{max}^y=1$). Right: VS-CMDE ($\sigma_{max}^{y(\text{target})}=1$).}
    \label{fig:VR}
\end{figure*}

Apart from \emph{continuous-time training} (where $t$ is sampled from $U(0,T)$), one can also perform \emph{discrete-time training} for the CMDE estimator. In discrete-time training, $t$ is sampled uniformly from $$\lbrace \epsilon, T/N, 2T/N,...,(N-1)T/N,T \rbrace.$$ We set $N=1000$, $T=1$ as in \cite{song2021sde}, $\sigma_{max}^y=1$ and trained the score model on the Edges2shoes dataset. We discovered that under this training configuration, the conditional score based model yielded a modelled distribution $p_{\theta}(x|y)$ (under reverse diffusion) that is far away from the true distribution $p_{\theta}(x|y)$, as shown visually in Figure \ref{fig:VR}. Apparently, the optimisation problem is harder because of the combination of discrete training and big difference in the diffusion speeds of $x$ and $y$. These factors can cause convergence to a bad local minimum.

For this reason, we decided to make the optimisation problem easier by gradually decreasing the diffusion rate of $y$ as the training proceeds. More specifically, the diffusion coefficient at training iteration $n$ of the VE SDE that governs the diffusion of the condition y is given by
\begin{equation}
    \begin{aligned}
\sigma^{(n)}(t) &= \sigma_{min} \left(\frac{\sigma_{\max}^{(n)}}{\sigma_{\min}}\right)^t, \text{with} \\ \sigma_{\max}^{(n)}&= \frac{M\sigma_{\max}^{\text{target}}\sigma_{\max}}{n(\sigma_{\max}-\sigma_{\max}^{\text{target}})+M\sigma_{\max}^{\text{target}}},
    \end{aligned}
    \label{eq:variance reduction}
\end{equation}
\noindent where $\sigma_{\max}^{\text{target}}$ is the target maximum standard deviation for the forward diffusion of $y$ at $M$ training iterations, and where $\sigma_{max}$ is the initial maximum standard deviation. Notice that $\sigma_{max}^{(n)}$ decreases in an inverse multiplicative rate and that $\sigma_{max}^{(M)}=\sigma_{\max}^{\text{target}}$. We dropped the superscript $y$ in eq. \ref{eq:variance reduction} for notational simplicity. We name this variation of CMDE as VS-CMDE. 

We performed the following experiment to test whether VS-CMDE outperforms CMDE: We set $M=125000$, $\sigma_{\max}^{\text{target}}=1$ and trained the score model for $125000$ iterations. At the end of training $\sigma_{max}^{y}$ reaches $1$, just as in the previous experiment, where the score model failed to reach a good approximation. However, training the score model with the variance reduction schedule led to a \emph{significantly} improved approximation of the posterior distribution as shown by the samples in Figure \ref{fig:VR}. The intuition behind the variance reduction schedule is that the problem of estimating $\nabla_{x_t, y_t}{\log{p(x_t, y_t)}}$ is easier if $y$ diffuses faster, since the distribution $p(x_t, y_t)$ is smoother, yielding an easier score approximation problem. The score model can fit the smooth score vector field more accurately at the early stage of the training. The initial fitting is used as a good initialisation for the harder task of estimating $\nabla_{x_t, y_t}{\log{p(x_t, y_t)}}$ at the later stages of training where y diffuses at a slower rate.

\begin{figure}
\renewcommand{\arraystretch}{1.25}
    \begin{tabular}{lccc}
        \begin{tabular}{@{}l@{}}
            Original image $x$
            \\[25pt]
        \end{tabular}
         & \includegraphics[width=.08\textwidth]{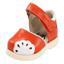} & \includegraphics[width=.08\textwidth]{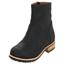} & \includegraphics[width=.08\textwidth]{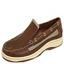} \\
        
        \begin{tabular}{@{}l@{}}
            Observation \\ $ y := Ax$
            \\[25pt]
        \end{tabular}
         & \includegraphics[width=.08\textwidth]{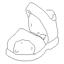} & \includegraphics[width=.08\textwidth]{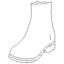} & \includegraphics[width=.08\textwidth]{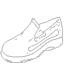} \\
        
         \begin{tabular}{@{}c@{}}
            Reconstruction $\hat{x}_1$
            \\[25pt]
        \end{tabular} & \includegraphics[width=.08\textwidth]{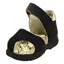}  & \includegraphics[width=.08\textwidth]{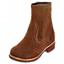}  & \includegraphics[width=.08\textwidth]{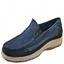} \\

        \begin{tabular}{@{}c@{}}
           Reconstruction $\hat{x}_2$
            \\[25pt]
        \end{tabular} & \includegraphics[width=.08\textwidth]{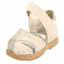} & \includegraphics[width=.08\textwidth]{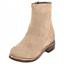} & \includegraphics[width=.08\textwidth]{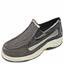}\\

        \begin{tabular}{@{}c@{}}
            Reconstruction $\hat{x}_3$
            \\[25pt]
        \end{tabular}  &  \includegraphics[width=.08\textwidth]{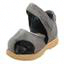} &  \includegraphics[width=.08\textwidth]{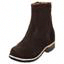} &  \includegraphics[width=.08\textwidth]{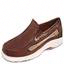} 
    \end{tabular}
    \caption{Results for the Edges2shoes task, using continuous-time training for CMDE.}
    \label{fig:CMDE continuous edges-to-shoes}
\end{figure}

When we switched to continuous training, CMDE performed well, as indicated by the visual samples in Figure \ref{fig:CMDE continuous edges-to-shoes} and the evaluation metrics presented in Table \ref{fig:edges-to-shoes}. However, VS-CMDE still outperformed vanilla CMDE. It is also worth mentioning that VS-CMDE yielded the same performance as CDE (in terms of JFID and LPIPS), while providing slightly more diverse samples. CDE outperformed CMDE based on the FID-scores in the Edges2shoes task, while performing on par with VS-CMDE. VS-CMDE performed on par with CMDE on the tasks of super-resolution and inpainting, as shown in Table \ref{tbl:augmented results}. 

Our experimental findings indicate, that in discrete training, VS-CMDE should be preferred over CMDE. However, in continuous training, VS-CMDE does not have a consistent competitive advantage over CMDE.

\begin{table*}
    \begin{center}
    \caption{Results of conditional generation tasks including VS-CMDE.}
    \label{tbl:augmented results}
    \begin{tabular}{cccccccc}
    \toprule
    &Estimator & PSNR/SSIM $\uparrow$  & LPIPS $\downarrow$ & UFID/JFID $\downarrow$ & Consistency $\uparrow$ & Diversity $\uparrow$ \\
    \midrule
    \multirow{3}{*}{Inpainting} 
    &CDE & \textbf{25.12}/\textbf{0.870}  & \textbf{0.042} & 13.07/18.06 & \textbf{28.54} & 4.79  \\
    &CDiffE & 23.07/0.844   & 0.057 & 13.28/19.25 &  26.61 & \textbf{6.52}   \\
    &CMDE ($\sigma^y_{max} = 1$) & 24.92/0.864  & 0.044 & \textbf{12.07/17.07} & 28.32 & 4.98  \\
    &VS-CMDE ($\sigma^{y(\text{target})}_{max} = 1$) & 24.87/0.863  & 0.044 & 12.14/17.21 & 28.37 & 5.08  \\
    \midrule
    \multirow{4}{*}{Super-resolution} 
    &CDE & 23.80/0.650  & 0.114 & 10.36/15.77 & 54.18 & \textbf{8.51}  \\
    &CDiffE & 23.83/0.656  & 0.139 & 14.29/20.20 & 51.90 & 7.41  \\
    &CMDE ($\sigma^y_{max} = 0.5$) & 23.91/0.654  & 0.109 & \textbf{10.28/15.68} & 53.03 & 8.33  \\
    &VS-CMDE ($\sigma^{y(\text{target})}_{max} = 0.5$) & 23.95/0.657 & 0.111 & 10.30/15.70 & 51.91 & 8.22 \\
    &HCFLOW & \textbf{24.95/0.702} & \textbf{0.107} & 14.13/19.55 & \textbf{55.31} & 6.26 \\
    \midrule
    \multirow{3}{*}{Edge to image} 
    &CDE & \textbf{18.35/0.699}  & \textbf{0.156} & \textbf{11.87/21.31} & \textbf{10.45} & 14.40 \\
    &CDiffE & 10.00/0.365   & 0.350 & 33.41/55.22  & 7.78 & \textbf{43.45} \\
    &CMDE ($\sigma^y_{max} = 1$) & 18.16/0.692  & 0.158 & 12.62/22.09 & 10.38 & 15.20  \\
    &VS-CMDE ($\sigma^{y(\text{target})}_{max} = 1$) & 18.28/0.695  & \textbf{0.156} & 12.16/\textbf{21.32} & 10.39 & 15.20  \\
    \bottomrule
    \end{tabular}
    \end{center}
\end{table*}

\section{Proofs}
\label{appendix:proofs}
\subsection{Equality of minimizers for CDE}
\label{appendix:minimizers}
\begin{lemma}
    \label{Vincent}
    For a fixed $y \in \mathbb{R}^d$ and $t \in \mathbb{R}$ we have
    \begin{align*}
        &\mathbb{E}_{\subalign{&x_0 \sim p(x_0 | y) \\ &x_t \sim p(x_t | x_0, y)}} 
            [\lambda(t) \norm{\nabla_{x_t} \ln{p(x_t | x_0, y)} - s_\theta(x_t, y, t)}_2^2]\\
        =& \mathbb{E}_{\subalign{&x_t \sim p(x_t |  y)}} 
            [\lambda(t) \norm{\nabla_{x_t} \ln{p(x_t | y)} - s_\theta(x_t, y, t)}_2^2]
    \end{align*}
\begin{proof}
    Since $y$ and $t$ are fixed, we may define $\psi(x_t) := s_\theta(x_t, y, t)$, $q(x_0) := p(x_0 | y)$ and $q(x_t | x_0) = p(x_t | x_0, y)$.
    Therefore, by the Tower Law, the statement of the lemma is equivalent to
    \begin{align*}
        &\mathbb{E}_{\subalign{&x_0, x_t \sim q(x_0, x_t)}} 
        [\norm{\nabla_{x_t} \ln{q(x_t | x_0)} - \psi(x_t)}_2^2]\\
    =& \mathbb{E}_{\subalign{&x_t \sim q(x_t)}} 
        [\norm{\nabla_{x_t} \ln{q(x_t)} - \psi(x_t)}_2^2]
    \end{align*}
    Which follows directly from \cite[Eq. 11]{vincent2011connection}.
\end{proof}
    
\end{lemma}
\begin{customthm}{1}
    The minimizer of
    \begin{gather*}
    \begin{aligned}
            &\frac{1}{2} \mathbb{E}_{\subalign{&t \sim U(0,T)\\ &x_0, y \sim p(x_0, y) \\ &x_t \sim p(x_t | x_0)}} 
            [\lambda(t) \norm{\nabla_{x_t} \ln{p(x_t | x_0)} - s_\theta(x_t, y, t)}_2^2]
    \end{aligned}
    \end{gather*}    
    in $\theta$ is the same as the minimizer of 
    \begin{gather*}
         \frac{1}{2} \mathbb{E}_{\subalign{&t \sim U(0,T)\\ &x_t, y \sim p(x_t, y)}} 
        [\lambda(t) \norm{\nabla_{x_t} \ln{p(x_t | y)} - s_\theta(x_t, y,t)}_2^2].
    \end{gather*}
\end{customthm}
\begin{proof}
    First, notice that $x_t$ is conditionally independent of $y$ given $x_0$. Therefore, by applying the Tower Law we obtain
    \begin{align*}
        &\mathbb{E}_{\subalign{&t \sim U(0,T)\\ &x_0, y \sim p(x_0, y) \\ &x_t \sim p(x_t | x_0)}} 
            [\lambda(t) \norm{\nabla_{x_t} \ln{p(x_t | x_0)} - s_\theta(x_t, y, t)}_2^2] \\
            \overset{(1)}{=}& \mathbb{E}_{\subalign{&t \sim U(0,T)\\ &y \sim p(y)\\&x_0 \sim p(x_0 | y) \\ &x_t \sim p(x_t | x_0)}} 
            [\lambda(t) \norm{\nabla_{x_t} \ln{p(x_t | x_0)} - s_\theta(x_t, y, t)}_2^2] \\
            \overset{(2)}{=}& \mathbb{E}_{\subalign{&t \sim U(0,T)\\ &y \sim p(y)\\&x_0 \sim p(x_0 | y) \\ &x_t \sim p(x_t | x_0, y)}} 
            [\lambda(t) \norm{\nabla_{x_t} \ln{p(x_t | x_0, y)} - s_\theta(x_t, y, t)}_2^2] \\ 
            =& \mathbb{E}_{\subalign{&t \sim U(0,T)\\ &y \sim p(y)}} 
            [f(t,y)] =: (*)
    \end{align*}
    where 
    \begin{align*}
        f&(t,y) := \\
        &\mathbb{E}_{\subalign{&x_0 \sim p(x_0 | y) \\ &x_t \sim p(x_t | x_0, y)}} [
        \lambda(t) \norm{\nabla_{x_t} \ln{p(x_t | x_0, y)} - s_\theta(x_t, y, t)}_2^2
    ].
    \end{align*}
    Now fix $y$ and $t$. By Lemma \ref{Vincent}, it follows that
    \begin{align*}
        &f(t,y) \\
        =&\mathbb{E}_{\subalign{&x_0 \sim p(x_0 | y) \\ &x_t \sim p(x_t | x_0, y)}} 
            [\lambda(t) \norm{\nabla_{x_t} \ln{p(x_t | x_0, y)} - s_\theta(x_t, y, t)}_2^2]\\
        \overset{(3)}{=}& \mathbb{E}_{\subalign{&x_t \sim p(x_t |  y)}} 
            [\lambda(t) \norm{\nabla_{x_t} \ln{p(x_t | y)} - s_\theta(x_t, y, t)}_2^2]\\
    \end{align*}
    Since $t$ and $y$ were arbitrary, this is true for all $t$ and $y$. Therefore, substituting back into $(*)$ we get that
    \begin{align*}
        (*) &= \mathbb{E}_{\subalign{&t \sim U(0,T)\\ &y \sim p(y)\\&x_t \sim p(x_t | y)}} 
            [\lambda(t) \norm{\nabla_{x_t} \ln{p(x_t | y)} - s_\theta(x_t, y, t)}_2^2]\\
        &\overset{(1)}{=} \mathbb{E}_{\subalign{&t \sim U(0,T)\\ &x_t, y \sim p(x_t, y)}} 
        [\lambda(t) \norm{\nabla_{x_t} \ln{p(x_t | y)} - s_\theta(x_t, y,t)}_2^2].
    \end{align*}
(1) Tower Law, (2) Conditional independence of $x_t$ and $y$ given $x_0$, (3) \text{Lemma} \ref{Vincent}.
\end{proof}

\subsection{Consistency of CDE}
\label{appendix:consistency}
In order to prove the consistency, in this subsection we make the following assumptions:
\begin{assumption}
    \label{assum:compact_1}
    The space of parameters $\Theta$ and the data space $\mathcal{X}$ are compact.
\end{assumption}
\begin{assumption}
    \label{assum:unique}
    There exists a unique $\theta^\ast \in \Theta$ such that $s_{\theta^\ast}(x,y,t) = \nabla_{x_t}\ln p (x,y,t)$.
\end{assumption}

First we state some technical, but well-known lemmas, which will be useful in proving our consistency result.

\begin{lemma}[Uniform law of large numbers] \cite[Lemma 2.4]{whitney1994estimation} \\
    \label{lemma:ULLN}
    Let $z_i$ be i.i.d from a distribution $q(z)$ and suppose that:
    \begin{itemize}
        \item $\Theta$ is compact.
        \item $f(z,\theta)$ is continuous for all $\theta \in \Theta$ and almost all $z$.
        \item $f(\cdot, \theta)$ is a measurable function of $z$ for each $\theta$.
        \item There exists $d: \mathcal{Z} \xrightarrow{} \mathbb{R}$ such that $\mathbb{E}[d(z)]<\infty$ and $\norm{f(z,\theta)} \leq d(z)$ for each $\theta$.
    \end{itemize}
    Then $\mathbb{E}_{z}[f(z,\theta)]$ is continuous in $\theta$, and $\frac{1}{n}\sum_{i=1}^n f(z_i, \theta)$ converges to $\mathbb{E}_{z}[f(z,\theta)]$ uniformly in probability, i.e.:
    \begin{gather*}
        \sup_\theta \norm{\frac{1}{n}\sum_{i=1}^n f(z_i, \theta) - \mathbb{E}_{z}[f(z,\theta)]} \overset{P}{\to} 0
    \end{gather*}

\end{lemma}

\begin{lemma}[Consistency of extremum estimators] \cite[Theorem 2.1]{whitney1994estimation} \\
    \label{lemma:consistency}
    Let $\Theta$ be compact and consider a family of functions $\mathcal{L}^{(n)}: \Theta \to \mathbb{R}$. Moreover, suppose there exists a function $\mathcal{L}: \Theta \to \mathbb{R}$ such that
    \begin{itemize}
        \item $\mathcal{L}(\theta)$ is uniquely minimized at $\theta^\ast$.
        \item $\mathcal{L}(\theta)$ is continuous.
        \item $\mathcal{L}^{(n)}(\theta)$ converges uniformly in probability to $\mathcal{L}(\theta)$.
    \end{itemize}
    Then $$\theta^\ast_n := \argmin_{\theta \in \Theta} \mathcal{L}^{(n)}(\theta) \overset{P}{\to} \theta^\ast.$$
\end{lemma}

\begin{customcoll}{1}
    Let $\theta_n^\ast$ be a minimizer of a $n$-sample Monte Carlo approximation of \begin{gather*}
        \begin{aligned}
                \frac{1}{2} \mathbb{E}_{\subalign{&t \sim U(0,T)\\ &x_0, y \sim p(x_0, y) \\ &x_t \sim p(x_t | x_0)}} 
                [\lambda(t) \norm{\nabla_{x_t} \ln{p(x_t | x_0)} - s_\theta(x_t, y, t)}_2^2].
        \end{aligned}
        \end{gather*} 
        Then under assumptions \ref{assum:compact_1} and \ref{assum:unique}, the conditional denoising estimator $s_{\theta_n^\ast}(x,y,t)$ is a consistent estimator of the conditional score $\nabla_{x_t} \ln p(x_t | y)$, i.e.
    \begin{gather*}
        s_{\theta_n^\ast}(x,y,t) \overset{P}{\to} \nabla_{x_t} \ln p(x_t | y),
    \end{gather*}
    as the number of Monte Carlo samples $n$ approaches infinity.
\end{customcoll}

\begin{proof}
    By conditional independence and the Tower Law, we get
    \begin{align*}
         &\mathbb{E}_{\subalign{&t \sim U(0,T)\\ &x_0, y \sim p(x_0, y) \\ &x_t \sim p(x_t | x_0)}} 
                [\lambda(t) \norm{\nabla_{x_t} \ln{p(x_t | x_0)} - s_\theta(x_t, y, t)}_2^2] \\
        =& \mathbb{E}_{\subalign{&t \sim U(0,T)\\ &x_0, y \sim p(x_0, y) \\ &x_t \sim p(x_t | x_0, y)}} 
            [\lambda(t) \norm{\nabla_{x_t} \ln{p(x_t | x_0)} - s_\theta(x_t, y, t)}_2^2] \\
        =& \mathbb{E}_{\subalign{&t \sim U(0,T)\\ &x_0,, x_t, y \sim p(x_0, x_t ,y)}} 
            [\lambda(t) \norm{\nabla_{x_t} \ln{p(x_t | x_0)} - s_\theta(x_t, y, t)}_2^2].
    \end{align*}
    Let $z=(t,x_0,x_t,y)$ and denote by $q(z):=p(t,x_0,x_t,y)$ the joint distribution. Moreover, define $f(z,\theta) := \lambda(t) \norm{\nabla_{x_t} \ln{p(x_t | x_0)} - s_\theta(x_t, y, t)}_2^2$. Since $t \sim U(0,T)$ is independent of $(x_0, x_t ,y) \sim p(x_0, x_t ,y)$, the above is equal to 
    \begin{align*}
        \mathbb{E}_{z \sim q(z)} 
            [f(z,\theta)]
    \end{align*}
    Therefore by Lemma \ref{lemma:ULLN}, the Monte Carlo approximation of \ref{CDN}: $\mathcal{L}^{(n)}(\theta)=\frac{1}{n}\sum_{i=1}^n f(z_i, \theta)$ converges uniformly in probability to $\mathcal{L}(\theta) = \mathbb{E}_{z \sim q(z)} 
    [f(z,\theta)]$. Let $\theta^\ast$ be the minimizer of $\mathcal{L}(\theta)$, by Lemma \ref{lemma:consistency} we get that $\theta^\ast_n \overset{P}{\to} \theta^\ast$ . Finally by Theorem \ref{thm:CDE_consistency}, $\theta^\ast$ is also a minimizer of the Fisher divergence between $s_{\theta^\ast}(x_t,y,t)$ and $\nabla_{x_t} \ln p(x_t | y)$ and by Assumption \ref{assum:unique} this implies that $s_{\theta^\ast}(x_t,y,t) = \nabla_{x_t} \ln p(x_t | y)$. Hence $s_{\theta_n^\ast}(x,y,t) \overset{P}{\to} \nabla_{x_t} \ln p(x_t | y)$.
\end{proof}

\subsection{Likelihood weighting for multi-speed and multi-sde models}
\label{appendix:weighting}
In this section we derive the likelihood weighting for multi-sde models (Theorem \ref{thm:weightning}). First using the framework in \cite[Appendix A]{song2021sde} we present the Anderson's theorem for multi-dimensional SDEs with non-homogeneous covariance matrix (without assuming $\Sigma(t) \not = \sigma(t) I$) and generalize the main result of \cite{song2021maximum} to this setting. Then, we cast the problem of multi-speed and  multi-sde diffusion as a special case of multi-dimensional diffusion with a particular covariance matrix $\Sigma(t)$ and thus obtain the likelihood weighting for multi-sde models (Theorem \ref{thm:weightning}).

Consider an Ito's SDE
\begin{align*}
    dx = \mu(x, t)dt + \Sigma(t)dw
\end{align*}
where $\mu: \mathbb{R}^{n_x} \times [0,T] \xrightarrow{} \mathbb{R}^{n_x}$ and $\Sigma: [0,T] \xrightarrow{} \mathbb{R}^{n_x \times n_x}$ is a time-dependent positive-definite matrix. By multi-dimensional Anderson's Theorem \cite{anderson1982reverse_time_sde} the corresponding reverse time SDE is given by 
\begin{align}
    \label{eq:true_rtsde}
    dx &= \tilde{\mu}(x, t)dt + \Sigma(t)dw \\
    \text{where } \tilde{\mu}(x,t) &:= \mu(x,t) -  \Sigma(t)^2 \nabla_x \ln p_{X_t}(x). \nonumber
\end{align}

If we train a score-based diffusion model to approximate $\nabla_x \ln p_{X_t}(x)$ with a neural network $s_\theta(x,t)$ we will obtain the following approximate reverse-time sde
\begin{align}
    \label{eq:approx_rtsde}
    dx &= \tilde{\mu}_\theta(x, t)dt + \Sigma(t)dw \\
    \text{where } \tilde{\mu}_\theta(x,t) &:= \mu(x,t) -  \Sigma(t)^2 s_\theta(x,t) \nonumber
\end{align}

Now we generalize \cite[Theorem 1]{song2021maximum} to multi-dimensional setting.
\begin{theorem}
    Let $p(x_t)$ and $p_\theta(x_t)$ denote marginal distributions of \ref{eq:true_rtsde} and \ref{eq:approx_rtsde} respectively. Then under regularity assumptions of  \cite[Theorem 1]{song2021maximum} we have that 
    \label{thm:multi-dim}
    \begin{align*}
        KL(p(x_0) | p_\theta(x_0)) \leq &KL(p(x_T) | \pi(x_T)) 
        \\ &+ \frac{1}{2} \mathbb{E}_{\subalign{&t \sim U(0,T)\\ &x_t \sim p(x_t)}} 
        [
            v^T \Sigma(t)^2 v
        ],
    \end{align*}
where $v=\nabla_{x_t} \ln{p(x_t)} - s_\theta(x_t,t)$. 
\end{theorem}
\begin{proof}   
    We proceed in close analogy to the proof of \cite[Theorem 1]{song2021maximum} but we use a more general diffusion matrix $\Sigma(t)$. 
    Let $P$ be the law of the true reverse-time sde and let $P_\theta$ be the law of the approximate reverse-time sde. 
    Then by  \cite[Theorem 2.4]{leonard2013properties} (generalized chain rule for KL divergence) we have
    \begin{align*}
        KL(P | P_\theta) = &KL(p(x_0) | p_\theta(x_0)) 
        \\ &+  \mathbb{E}_{z \sim p(x_0)}[KL(P(\cdot | x_0=z) |P_\theta(\cdot | x_0=z))].
    \end{align*}
    Since $\mathbb{E}_{z \sim p(x_0)}[KL(P(\cdot | x_0=z) |P_\theta(\cdot | x_0=z))] \geq 0$, this implies 
    \begin{align*}
       KL(p(x_0) | p_\theta(x_0)) \leq KL(P | P_\theta) 
    \end{align*}
    Using the fact that $p_\theta(x_T) = \pi$ and applying \cite[Theorem 2.4]{leonard2013properties} again, we obtain
    \begin{align*}
        KL(P | P_\theta) =  &KL(p(x_T) |\pi) 
        \\ &+  \mathbb{E}_{\subalign{z \sim p(x_T)}}[KL(P(\cdot | x_T=z) |P_\theta(\cdot | x_T=z))].
    \end{align*}
    Let $P^z := P(\cdot | x_T=z)$ and $P_\theta^z := P_\theta(\cdot | x_T=z)$
    \begin{align*}
        \mathbb{E}_{\subalign{z \sim p(x_T)}}[KL(P(\cdot | x_T=z) |P_\theta(\cdot | x_T=z))]
        \\ =  - \mathbb{E}_{\subalign{z \sim p(x_T)}} \left[ 
            \mathbb{E}_{P^z} \left[
                \ln \frac{d P^z_\theta}{d P^z}
            \right]
        \right]
    \end{align*}
    Using Girsanov Theorem \cite[Theorem 8.6.5]{oksendal2003sde} and the fact that $\Sigma(t)$ is symmetric and invertible 
    \begin{align*}
     =  &\mathbb{E}_{z \sim p(x_T)}  \bigg[ 
            \mathbb{E}_{P^z} \bigg[\\
                    &\int_0^T  \Sigma(t) v(x_t,t) dw_t 
                    + \frac{1}{2} \int_0^T v(x_t,t)^T \Sigma(t)^2 v(x_t,t) dt 
            \bigg]
        \bigg]
    \end{align*}
    where $v(x_t,t)=\nabla_{x_t} \ln{p(x_t)} - s_\theta(x_t,t)$. Since $\int_0^T  \Sigma(t) v(x_t,t) dw_t $ is a martingale (Ito's integral wrt Brownian motion) 
    \begin{align*}
        &=  \frac{1}{2} \mathbb{E}_{z \sim p(x_T)}  \bigg[ 
            \mathbb{E}_{P^z} \bigg[
                     \int_0^T v(x_t,t)^T \Sigma(t)^2 v(x_t,t) dt 
            \bigg]
        \bigg] \\
        &= \frac{1}{2} \int_0^T \mathbb{E}_{x \sim p(x_t)}[ v(x_t,t)^T \Sigma(t)^2 v(x_t,t)] \\
        &= \frac{1}{2} \mathbb{E}_{\subalign{&t \sim U(0,T)\\ &x_t \sim p(x_t)}} 
        [
            v(x_t,t)^T \Sigma(t)^2 v(x_t,t)
        ].
    \end{align*}
\end{proof}

\subsubsection{Multi-sde and multi-speed diffusion}
Now we consider again the multi-speed and the more general multi-sde diffusion frameworks from Sections \ref{sec:CMDE} and \ref{sec:multi-sde}. Suppose that we have two tensors $x$ and $y$ which diffuse according to different SDEs
\begin{gather*}
    dx = \mu^x(x,t)dt+\sigma^x(t)dw  \\
    dy = \mu^y(y,t)dt+\sigma^y(t)dw  
\end{gather*}
We may cast this system of two SDEs, as a single SDE
\begin{gather*}
    dz = \mu^z(z,t)dt+ \Sigma_z(t)dw 
\end{gather*}
where $z = (x,y)$, $\mu^z(z,t) = (\mu^x(x,t), \mu^y(x,t))$ and 
\begin{gather*}
    \Sigma_z(t) =  
    \begin{cases} 
        \sigma^x(t), \text{ if } i=j, \ i \leq n_x \\ 
        \sigma^y(t), \text{ if } i=j, \ n_x < i \leq n_y  \\
        0, \text{ otherwise}
    \end{cases}
    .            
\end{gather*}
If we train a score-based diffusion model for $z_t = (x_t, y_t)$, then by Theorem \ref{thm:multi-dim}
\begin{align*}
    KL(p(z_0) | p_\theta(z_0)) \leq C_1 + \frac{1}{2} \mathbb{E}_{\subalign{&t \sim U(0,T)\\ &z_t \sim p(z_t)}} 
    [
        v^T \Sigma_z(t)^2 v
    ],
\end{align*}
where $C_1 := KL(p(x_T) | \pi(x_T)) $ does not depend on $\theta$.
Because $\Lambda_{MLE}$ (from Theorem \ref{thm:weightning}) is equal to $\Sigma_z(t)^2$, we may rewrite the above as 
\begin{align*}
    KL(p(z_0) | p_\theta(z_0)) \leq C_1 + \frac{1}{2} \mathbb{E}_{\subalign{&t \sim U(0,T)\\ &z_t \sim p(z_t)}} 
    [
        v^T \Lambda_{MLE}(t)^2 v
    ],
\end{align*}
and since by denoising score matching \cite{vincent2011connection} 
\begin{align*}
    \mathbb{E}_{\subalign{&t \sim U(0,T)\\ &z_t \sim p(z_t)}} 
[
    v^T \Lambda_{MLE}(t) v
] &= 
\\ &\mathbb{E}_{\subalign{&t \sim U(0,T)\\ &z_0 \sim p_0(z_0) \\ &z_t \sim p(z_t | z_0)}} 
[
    v^T \Lambda_{MLE}(t) v
] + C_2
\end{align*}
where $C_2$ is another term constant in $\theta$.
We conclude that 
\begin{align*}
    KL(p(z_0) | p_\theta(z_0)) \leq \frac{1}{2} \mathbb{E}_{\subalign{&t \sim U(0,T)\\ &z_0 \sim p_0(z_0) \\ &z_t \sim p(z_t | z_0)}} 
    [
        v^T \Lambda_{MLE}(t) v
    ] + C_3
\end{align*}
where $C_3 := C_1 + C_2$.
Now recall that the term on the RHS is exactly the training objective of a multi-sde score-based diffusion model with likelihood weighting
\begin{gather*}
    \begin{aligned}
        \mathcal{L}(\theta) := \frac{1}{2} \mathbb{E}_{\subalign{&t \sim U(0,T)\\ &z_0 \sim p_0(z_0) \\ &z_t \sim p(z_t | z_0)}} 
        [
            v^T \Lambda_{MLE}(t) v
        ].
    \end{aligned}
\end{gather*}
Therefore
\begin{align*}
    KL(p(z_0) | p_\theta(z_0)) \leq \mathcal{L}(\theta)  + C_3.
\end{align*}
Finally, since $KL(p(z_0) | p_\theta(z_0))  = \mathbb{E}_{(x,y) \sim p(x,y)}[\ln p(x,y)]  - \mathbb{E}_{(x,y) \sim p(x,y)}[\ln p_\theta(x,y)]$, we have 
\begin{gather*}
    -\mathbb{E}_{(x,y) \sim p(x,y)}[\ln p_\theta(x,y)] \leq \mathcal{L}(\theta) + C
\end{gather*}
where $C := C_3 - \mathbb{E}_{(x,y) \sim p(x,y)}[\ln p(x,y)] $ is independent of $\theta$. Thus the Theorem \ref{thm:weightning} is established.

\subsection{Mean square approximation error}
\label{appendix:mse}

\begin{assumption}
    \label{assum: c2}
    $p(x,y) \in C^2(\mathcal{X})$\
\end{assumption}
\begin{assumption}
    \label{assum: lower_bound}
    $p(x,y) > 0$ for all $x,y$.
\end{assumption}
\begin{assumption}
    \label{assum: compact_2}
    The data space $\mathcal{X}$ is compact.
\end{assumption}

\begin{lemma}
    \label{lemma: blurring}
    Under assumptions \ref{assum: c2} and \ref{assum: compact_2} we have
    \begin{gather*}
        p_{Y_t | X_t}(y_t | x_t)  = (p_{Y | X_t}(\cdot | x_t) \ast \varphi_\sigma)(y_t) \\
        \partial_{x_t} p_{Y_t | X_t}(y_t | x_t) = (\partial_{x_t} p_{Y| X_t}(\cdot | x_t) \ast \varphi_\sigma)(y_t)
    \end{gather*}
\end{lemma}
\begin{proof}
    For this proof, we drop our convention of denoting the probability distribution of a random variable via the name of its density’s argument.
    \begin{align*}
        p_{Y_t | X_t}(y_t | x_t)  &=
        \\ &= \int p_{Y, Y_t| X_t}(y, y_t | x_t) dy 
        \\ &= \int p_{Y | X_t}(y | x_t) p_{Y_t |Y, X_t}(y_t | y, x_t) dt
        \\ &= \int p_{Y | X_t}(y | x_t) p_{Y_t |Y}(y_t | y) dy
    \end{align*}
    Since $Y_t |Y$ has normal distribution with mean $y$ and variance $\sigma^y(t)^2$:
    \begin{align*}
        &= \int p_{Y | X_t}(y | x_t) \varphi_\sigma(y_t - y)  dy 
        \\ &= (p_{Y | X_t}(\cdot | x_t) \ast \varphi_\sigma)(y_t)
    \end{align*}
    where $\varphi_\sigma$ is a Gaussian kernel with variance $\sigma^y(t)^2$.
    Moreover, under the assumptions of the lemma we can exchange the differentiaion and integration. Therefore
    \begin{align*}
        \partial_{x_t} p_{Y_t | X_t}(y_t | x_t) &= \partial_{x_t} \int p_{Y | X_t}(y | x_t) \varphi_\sigma(y_t - y) dy   
        \\ &=  \int \partial_{x_t} p_{Y | X_t}(y | x_t) \varphi_\sigma(y_t - y) dy  
        \\ &= (\partial_{x_t} p_{Y | X_t}(\cdot | x_t) \ast \varphi_\sigma)(y_t)
    \end{align*}
\end{proof}

\begin{lemma}
    \label{lemma: sup_norm}
    Let $f$ be a $C^1$-function on a compact domain $\mathcal{X}$ and let $\varphi_\sigma$ be a Gaussian kernel with variance $\sigma^2$. Then there exists a function $E: \mathbb{R} \xrightarrow{} \mathbb{R}$, which is monotonically decreasing to zero, such that
    \begin{gather*}
        \norm{(f \ast \varphi_\sigma) - f}_{\infty} \leq E(1/\sigma).
    \end{gather*}
\end{lemma}

\begin{proof}
    \begin{align*}
        & \phantom{=}|(f \ast \varphi_\sigma)(y) -f(y)| 
        \\&= \bigg| \int f(z)  \varphi_\sigma(z-y) dz - \int f(y) \varphi_\sigma(z-y) dz \bigg|
        \\&\leq  \int |f(z) -f(y)| \varphi_\sigma(z-y) dz
    \end{align*}
    Since $f$ is  a $C^1$ function on a compact domain, it is Lipschitz and bounded (in absolute value) by some constant $M$. 
    Fix $\epsilon > 0$, and let $L$ denote the Lipschitz constant of $f$.
    We have that $|f(z) -f(y)| < \epsilon$ whenever $\norm{z-y} < \epsilon / L$.
    Let $B_y( \epsilon / L ) := \{z \in \mathcal{X} : \norm{z-y} < \epsilon / L \}$ be a ball of radius $\epsilon / L$ around $y$.
    Then
    \begin{align*}
        &\phantom{=}\int |f(z) -f(y)| \varphi_\sigma(z-y) dz 
        \\ &\! \begin{aligned}
            = \int_{B_y( \epsilon / L )}& |f(z) -f(y)| \varphi_\sigma(z-y) dz 
            \\ &+ \int_{\mathcal{X} \setminus B_y( \epsilon / L )} |f(z) -f(y)| \varphi_\sigma(z-y) dz
        \end{aligned}
        \\ &\leq \epsilon +  \int_{\mathcal{X} \setminus B_y( \epsilon / L )}  2M \varphi_\sigma(z-y) dz
       \\ &= \epsilon +  2M P \left( |Z_\sigma| > \frac{\epsilon}{L} \right)
    \end{align*}
    where $Z_\sigma$ is a normally-distributed random variable with mean zero and variance $\sigma^2$.
    By the Chernoff bound, we have
    \begin{align*}
        \leq  \epsilon +  4M \exp \left( -\frac{\epsilon^2}{2L^2 \sigma^2}\right).
    \end{align*}
    
    \noindent Define $E_\epsilon(1/\sigma) :=   \epsilon +  4M \exp \left( -\frac{\epsilon^2}{2L^2 \sigma^2}\right)$. Observe that $E_\epsilon: \mathbb{R}_+ \xrightarrow{} \mathbb{R}$ is monotonically decreasing to $\epsilon$.
    Moreover 
    $$ \norm{(f \ast \varphi_\sigma) - f}_{\infty} \leq E_{\epsilon}(1/\sigma). $$
    Now let $A := [0,1]$ and define 
    $$E(1/\sigma) := \min_{\epsilon \in A} E_\epsilon(1/\sigma).$$
    Notice that the above minimum is achieved, since $A$ is compact and for a fixed $\sigma$, the function $\epsilon \mapsto E_{\epsilon}(1/\sigma)$ is continuous.

    We will prove that $E$ is a monotonically decreasing to zero and upper-bounds $\norm{(f \ast \varphi_\sigma) - f}_{\infty}$.
    Firstly, it is clear that $E(x) \to 0$ as $x \to \infty$, since for all $\epsilon \in A$ we have $\lim_{x \to \infty} E(x) \leq \lim_{x \to \infty} E_{\epsilon}(x)=\epsilon$ . 
    Secondly, suppose $a < b$, and let $\epsilon_a$ be such that $E(a) = E_{\epsilon_a}(a)$. Then
    $$E(b) = \inf_{\epsilon \in A} E_{\epsilon}(b) \leq E_{\epsilon_a}(b) < E_{\epsilon_a}(a) = E(a).$$
    Therefore $E$ is monotonically decreasing.
    Finally since for all $\epsilon > 0$
    $$ \norm{(f \ast \varphi_\sigma) - f}_{\infty} \leq E_{\epsilon}(1/\sigma). $$
    Taking minimum over $\epsilon \in A$ on both sides we obtain 
    $$ \norm{(f \ast \varphi_\sigma) - f}_{\infty} \leq E(1/\sigma). $$
\end{proof}

\begin{lemma}
    \label{lemma: Lpz}
    Let $f$ be a $C^1$ function on a compact domain and let $Z$ be a random variable with mean $\mu$ and variance $\sigma^2$.
    Then
    \begin{gather*}
        \mathbb{E}_Z[(f(\mu) - f(Z))^2] \leq L^2 \sigma^2
    \end{gather*}
    where $L$ denotes the Lipschitz constant of $f$.
\end{lemma}
\begin{proof}
Since  $f$ is a $C^1$ function on a compact domain it is Lipschitz with some Lipschitz constant $L$. Therefore
\begin{align*}
    \mathbb{E}_Z[(f(\mu) - f(Z))^2] 
    \leq L^2\mathbb{E}_Z[(\mu - Z)^2]
    \leq L^2 \sigma^2
\end{align*}    
\end{proof}

\begin{customthm}{3}
    Fix $t$, $x_t$ and $y$. Then under Assumptions \ref{assum: c2}, \ref{assum: lower_bound} and \ref{assum: compact_2}, there exists a function $E: \mathbb{R} \xrightarrow{} \mathbb{R}$ which is monotonically decreasing to zero, such that
    \begin{gather*}
        \mathbb{E}_{y_t \sim p(y_t|y)}[
            \norm{ \nabla_{x_t} \ln p(x_t|y_t) - \nabla_{x_t} \ln p(x_t|y)}_2^2
            ] \\
            \leq E(1/\sigma^y(t)).
    \end{gather*}
\end{customthm}
\begin{proof}
    For this proof, we drop our convention of denoting the probability distribution of a random variable via the name of its density’s argument.
    \begin{align*}
        \norm{ \nabla_{x_t} \ln p_{X_t | Y_t}(x_t|y_t) - \nabla_{x_t} \ln p_{X_t | Y}(x_t|y)}_2^2
    \\= \sum_{i=1}^{n_x} ( \partial^i_{x_t} \ln p_{X_t | Y_t}(x_t|y_t) - \partial^i_{x_t} \ln p_{X_t | Y}(x_t|y) )^2
    \end{align*}
    Therefore it is sufficient to prove the theorem in each dimension separately. Hence, without loss of generality, we may assume that $x_t \in \mathbb{R}$ and show
    \begin{gather*}
        \mathbb{E}_{y_t \sim p(y_t|y)}[
            ( \partial_{x_t} \ln p_{X_t | Y_t}(x_t|y_t) - \partial_{x_t} \ln p_{X_t | Y}(x_t|y) )^2
        ]  
        \\ \leq E(1/\sigma^y(t)).
    \end{gather*}
    By Bayes's rule we have
    \begin{align*}
        &\partial_{x_t} \ln p_{X_t | Y_t}(x_t|y_t)  =  \partial_{x_t} \ln p_{Y_t | X_t}(y_t | x_t) + \partial_{x_t} \ln p_{X_t}(x_t)
        \\ &\partial_{x_t} \ln  p_{X_t | Y}(x_t|y)  = \partial_{x_t} \ln p_{Y | X_t}(y | x_t) + \partial_{x_t} \ln p_{X_t}(x_t).
    \end{align*}
    Therefore,
    \begin{align*}
        &(  \partial_{x_t} \ln p_{X_t | Y_t}(x_t|y_t)  - \partial_{x_t} \ln p_{X_t | Y}(x_t|y) )^2
        \\ &= (\partial_{x_t} \ln p_{Y_t | X_t}(y_t | x_t)- \partial_{x_t} \ln p_{Y | X_t}(y | x_t) )^2.
    \end{align*}
    To unclutter the notation, let $p(y | x) := p_{Y | X_t}(y | x)$ and $p_\sigma(y | x) :=  p_{Y_t | X_t}(y | x)$. Applying this notation:
    \begin{align*}
        \mathbb{E}_{y_t \sim p(y_t|y)}[
            (\partial_{x_t} \ln p_{Y_t | X_t}(y_t | x_t)- \partial_{x_t} \ln p_{Y | X_t}(y | x_t) )^2] \\
        = 
        \mathbb{E}_{y_t \sim p(y_t|y)}[
            (\partial_{x_t} \ln p_\sigma(y_t | x_t) - \partial_{x_t}  \ln p(y | x_t) )^2 ]
    \end{align*}
    Adding and subtracting $\partial_{x_t} \ln p(y_t | x_t)$ and using the triangle inequality:
    \begin{align*}
        \! \begin{aligned} 
            \leq &\mathbb{E}_{y_t \sim p(y_t|y)}[
                (\partial_{x_t} \ln p_\sigma(y_t | x_t) - \partial_{x_t} \ln p(y_t | x_t) )^2] \\
            &+ \mathbb{E}_{y_t \sim p(y_t|y)}[
                (\partial_{x_t} \ln p(y_t | x_t)- \partial_{x_t} \ln p(y | x_t) )^2]
        \end{aligned}
    \end{align*}
    We may bound the expectation by the supremum norm
    \begin{align*}
        \! \begin{aligned} 
            \leq & \norm{\partial_{x_t} \ln p_\sigma( \cdot | x_t) - \partial_{x_t} \ln p( \cdot | x_t) }_{\infty}^2 \\
            &+ \mathbb{E}_{y_t \sim p(y_t|y)}[(\partial_{x_t} \ln p(y_t | x_t)- \partial_{x_t} \ln p(y | x_t) )^2]
        \end{aligned}
    \end{align*}
    We will bound each of the summands separately. Firstly, by Assumption \ref{assum: c2}  $(y_t, x_t) \to p(y_t | x_t)$ is $C^2$ and therefore $(y_t, x_t) \to \partial_{x_t}p(y_t | x_t)$ is $C^1$. Moreover, since $\mathcal{X}$ is compact,  $y_t \to \partial_{x_t}p(y_t | x_t)$  is Lipschitz for some Lipschitz constant $L$. 
    Therefore, by Lemma \ref{lemma: Lpz},
    \begin{align*}
        \mathbb{E}_{y_t \sim p(y_t|y)}[ (\partial_{x_t} \ln p(y_t | x_t)- \partial_{x_t} \ln p(y | x_t) )^2] \leq  L^2 \sigma^y(t)^2.
    \end{align*} 
    To finish the proof, we need to bound $$ \norm{\partial_{x_t} \ln p_\sigma( \cdot | x_t) - \partial_{x_t} \ln p( \cdot | x_t) }_{\infty}^2 $$
    First, we apply the chain rule
    \begin{align*}
        &\phantom{=}\norm{\partial_{x_t} \ln p_\sigma( \cdot | x_t) - \partial_{x_t} \ln p( \cdot | x_t) }_{\infty}^2 
        \\ &=  \norm{ 
            \frac{\partial_{x_t} p_{\sigma}( \cdot | x_t)}{ p_{\sigma}( \cdot | x_t)}  
            - \frac{\partial_{x_t} p( \cdot | x_t)}{ p( \cdot | x_t)} 
        }_{\infty}^2 
    \end{align*}
    Adding and subtracting $\frac{\partial_{x_t} p_{\sigma}( \cdot | x_t)}{ p( \cdot | x_t)} $:
    \begin{align*}
        \ \! &\begin{aligned}
             \leq &\norm{                  
            \frac{\partial_{x_t} p_{\sigma}( \cdot | x_t)}{ p_{\sigma}( \cdot | x_t)}  
            - \frac{\partial_{x_t} p_{\sigma}( \cdot | x_t)}{ p( \cdot | x_t)} 
        }_{\infty}^2 
        \\ &+  \norm{ 
            \frac{\partial_{x_t} p_{\sigma}( \cdot | x_t)}{ p( \cdot | x_t)}  
            - \frac{\partial_{x_t} p( \cdot | x_t)}{ p( \cdot | x_t)} 
        }_{\infty}^2 
        \end{aligned}
        \\ \! &\begin{aligned}
             = &\norm{ 
            \frac{\partial_{x_t} p_{\sigma}( \cdot | x_t)[ p( \cdot | x_t) - p_{\sigma}( \cdot | x_t)]}
            { p_{\sigma}( \cdot | x_t) p( \cdot | x_t)}  
        }_{\infty}^2 
        \\ &+  \norm{ 
            \frac{\partial_{x_t} p_{\sigma}( \cdot | x_t) - \partial_{x_t} p( \cdot | x_t)}
            { p( \cdot | x_t)}  
        }_{\infty}^2 
        \end{aligned}
    \end{align*}
    By assumption \ref{assum: c2} and \ref{assum: compact_2} we have that $\partial_{x_t} p_{\sigma}( \cdot | x_t)$, $p_{\sigma}( \cdot | x_t)$ and  $p( \cdot | x_t)$ are continuous functions on a compact domain. Therefore, $\partial_{x_t} p_{\sigma}( \cdot | x_t)$ is bounded from above by some constant $M$. Moreover, by adding assumption \ref{assum: lower_bound} we obtain that $p_{\sigma}( \cdot | x_t)$ and  $p( \cdot | x_t)$ are bounded from below by some $\epsilon > 0$. Therefore 
    \begin{align*}
        \! &\begin{aligned}
            \leq &\norm{ 
           \frac{\partial_{x_t} p_{\sigma}( \cdot | x_t)[ p( \cdot | x_t) - p_{\sigma}( \cdot | x_t)]}
           { p_{\sigma}( \cdot | x_t) p( \cdot | x_t)}  
       }_{\infty}^2 
       \\ &+  \norm{ 
           \frac{\partial_{x_t} p_{\sigma}( \cdot | x_t) - \partial_{x_t} p( \cdot | x_t)}
           { p( \cdot | x_t)}  
       }_{\infty}^2 
       \end{aligned}
       \\ \! &\begin{aligned}
        \leq \frac{M}{\epsilon^2} &\norm{ p( \cdot | x_t) - p_{\sigma}( \cdot | x_t)
   }_{\infty}^2 
   \\ &+ \frac{1}{\epsilon}  \norm{ \partial_{x_t} p_{\sigma}( \cdot | x_t) - \partial_{x_t} p( \cdot | x_t)
   }_{\infty}^2 
   \end{aligned}
    \end{align*}
    Now by Lemma \ref{lemma: sup_norm} and \ref{lemma: blurring}
    \begin{align*}
        \leq \frac{M}{\epsilon^2} E_1(1/\sigma^y(t)^2) + \frac{1}{\epsilon}  E_2(1/\sigma^y(t)^2)
    \end{align*}
    where $E_1$ and $E_2$ are monotonically decreasing to zero.
    The theorem follows with $E(1/\sigma^y(t)^2) := \frac{M}{\epsilon^2} E_1(1/\sigma^y(t)^2) + \frac{1}{\epsilon}  E_2(1/\sigma^y(t)^2) + L^2 \sigma^y(t)^2$, which monotonically decreases to zero as $\sigma^y(t)^2$ decreases to zero.
\end{proof}

\section{Architectures and hyperparameters}

We used almost the same neural network architecture across all tasks and all estimators, so that we can compare the estimators fairly. The only difference between the score model for the diffusive estimators and the score model for the CDE estimator is that the former contains $6$ instead $3$ filters in the final convolution to account for the joint score estimation. This difference in the final convolution leads to negligible difference in the number of parameters, which is highly unlikely to have impacted the final performance. 

We used the basic version of the DDPM architecture with the following hyperparameters: channel dimension $96$, depth multipliers $[1, 1, 2, 2, 3, 3]$, $2$ ResNet Blocks per scale and attention in the final $3$ scales. The total parameter count is 43.5M. Song et al. \cite{song2021sde} report improved performance with the NCSN++ architecture over the baseline DDPM when training with the VE SDE. This claim is also supported by the work of Saharia et al. \cite{saharia2021sr3}. Therefore, adopting this architecture is likely to improve the performance of all estimators and lead to even more competitive performance over state-of-the-art methods. For all estimators, we concatenate the condition image $y$ or $y(t)$ with the diffused target $x(t)$ and pass the concatenated image as input to the score model for score calculation. In the super-resolution experiment, we first interpolate the condition to the same resolution as the target using nearest neighbours interpolation and then concatenate it with the target image. 

We used exponential moving average (EMA) with rate 0.999 and the same optimizer settings as in \cite{song2021sde}. Moreover, we used a batch size of $50$ for the super-resolution and edge to image translation experiments and a batch size of $100$ for the inpainting experiments.

\section{Extended visual results}

\begin{figure*}
    \begin{center}
    \begingroup
    \setlength{\tabcolsep}{0pt}

    \begin{tabular}{ccccccc}
        Original image $x$ & Observation $ y$ & HCFlow & CDE & CDiffE & CMDE & VS-CMDE \\

        \includegraphics[width=.14\textwidth]{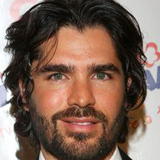} &   
        \includegraphics[width=.14\textwidth]{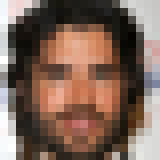} &
        \includegraphics[width=.14\textwidth]{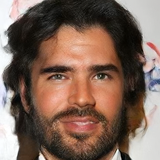} &
        \includegraphics[width=.14\textwidth]{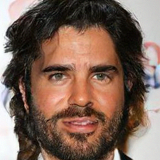} & 
        \includegraphics[width=.14\textwidth]{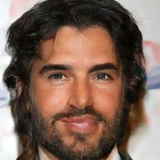} &
        \includegraphics[width=.14\textwidth]{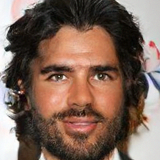} &
        \includegraphics[width=.14\textwidth]{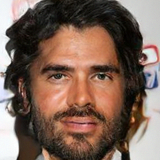}\\
        
        \includegraphics[width=.14\textwidth]{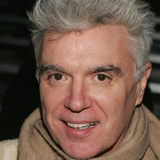} &   
        \includegraphics[width=.14\textwidth]{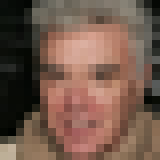} &
        \includegraphics[width=.14\textwidth]{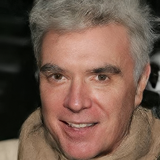} &
        \includegraphics[width=.14\textwidth]{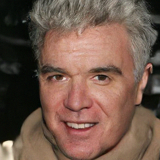} & 
        \includegraphics[width=.14\textwidth]{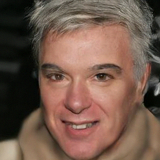} &
        \includegraphics[width=.14\textwidth]{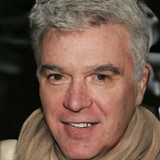} &
        \includegraphics[width=.14\textwidth]{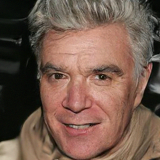}\\
        
        \includegraphics[width=.14\textwidth]{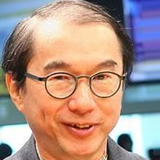} &   
        \includegraphics[width=.14\textwidth]{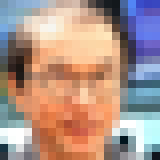} &
        \includegraphics[width=.14\textwidth]{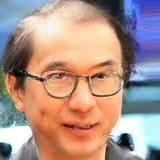} &
        \includegraphics[width=.14\textwidth]{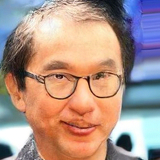} & 
        \includegraphics[width=.14\textwidth]{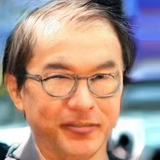} &
        \includegraphics[width=.14\textwidth]{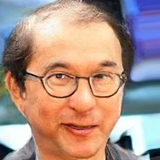} &
        \includegraphics[width=.14\textwidth]{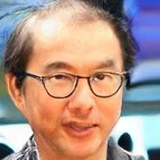}\\
        
        \includegraphics[width=.14\textwidth]{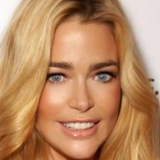} &   
        \includegraphics[width=.14\textwidth]{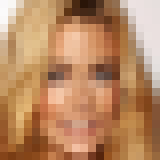} &
        \includegraphics[width=.14\textwidth]{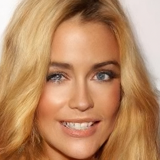} &
        \includegraphics[width=.14\textwidth]{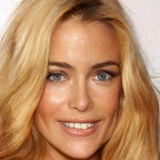} & 
        \includegraphics[width=.14\textwidth]{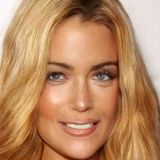} &
        \includegraphics[width=.14\textwidth]{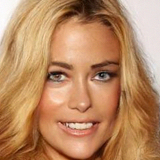} &
        \includegraphics[width=.14\textwidth]{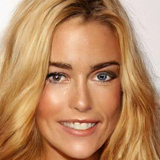}\\
        
        \includegraphics[width=.14\textwidth]{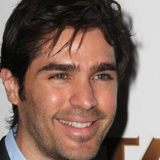} &   
        \includegraphics[width=.14\textwidth]{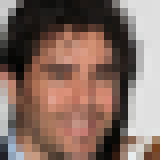} &
        \includegraphics[width=.14\textwidth]{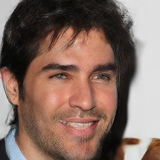} &
        \includegraphics[width=.14\textwidth]{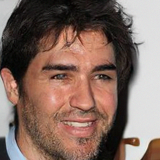} & 
        \includegraphics[width=.14\textwidth]{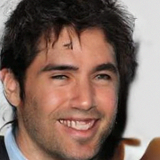} &
        \includegraphics[width=.14\textwidth]{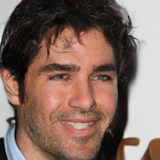} &
        \includegraphics[width=.14\textwidth]{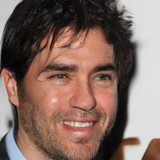}\\
        
        \includegraphics[width=.14\textwidth]{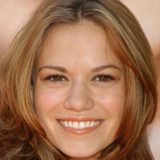} &   
        \includegraphics[width=.14\textwidth]{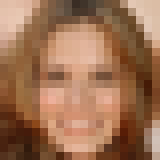} &
        \includegraphics[width=.14\textwidth]{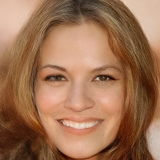} &
        \includegraphics[width=.14\textwidth]{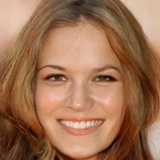} & 
        \includegraphics[width=.14\textwidth]{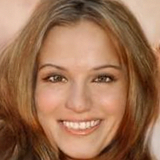} &
        \includegraphics[width=.14\textwidth]{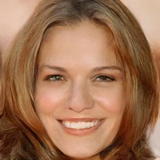} &
        \includegraphics[width=.14\textwidth]{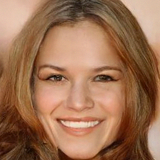}\\
        
        \includegraphics[width=.14\textwidth]{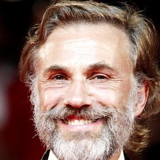} &   
        \includegraphics[width=.14\textwidth]{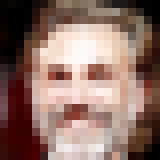} &
        \includegraphics[width=.14\textwidth]{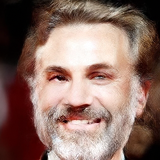} &
        \includegraphics[width=.14\textwidth]{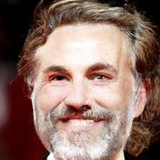} & 
        \includegraphics[width=.14\textwidth]{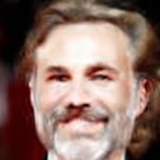} &
        \includegraphics[width=.14\textwidth]{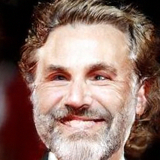} &
        \includegraphics[width=.14\textwidth]{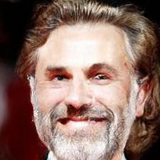}\\
        
        \includegraphics[width=.14\textwidth]{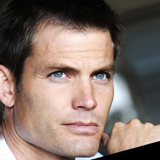} &   
        \includegraphics[width=.14\textwidth]{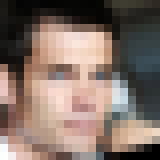} &
        \includegraphics[width=.14\textwidth]{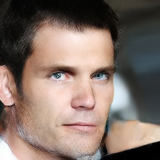} &
        \includegraphics[width=.14\textwidth]{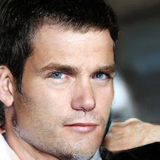} & 
        \includegraphics[width=.14\textwidth]{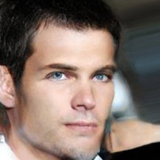} &
        \includegraphics[width=.14\textwidth]{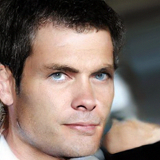} &
        \includegraphics[width=.14\textwidth]{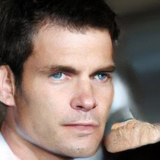}\\

    \end{tabular}
    \endgroup
    \end{center}
    \caption{Extended super-resolution results.}
    \label{fig:additional_sr}
\end{figure*}

\begin{figure*}
    \begin{center}
    \begingroup
    \setlength{\tabcolsep}{0pt}

    \begin{tabular}{ccccccc}
        Original image $x$ & Observation $ y$ & CDE & CDiffE & CMDE & VS-CMDE \\

        \includegraphics[width=.145\textwidth]{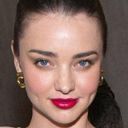} &   
        \includegraphics[width=.145\textwidth]{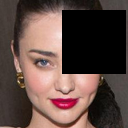} &
        \includegraphics[width=.145\textwidth]{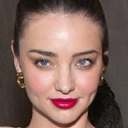} & 
        \includegraphics[width=.145\textwidth]{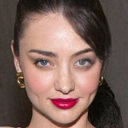} &
        \includegraphics[width=.145\textwidth]{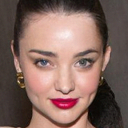} &
        \includegraphics[width=.145\textwidth]{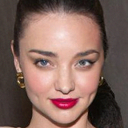}\\
        
        \includegraphics[width=.145\textwidth]{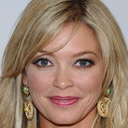} &   
        \includegraphics[width=.145\textwidth]{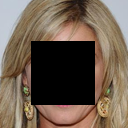} &
        \includegraphics[width=.145\textwidth]{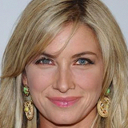} & 
        \includegraphics[width=.145\textwidth]{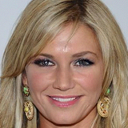} &
        \includegraphics[width=.145\textwidth]{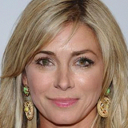} &
        \includegraphics[width=.145\textwidth]{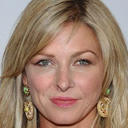}\\
        
        \includegraphics[width=.145\textwidth]{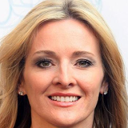} &   
        \includegraphics[width=.145\textwidth]{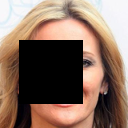} &
        \includegraphics[width=.145\textwidth]{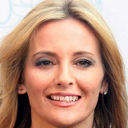} & 
        \includegraphics[width=.145\textwidth]{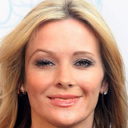} &
        \includegraphics[width=.145\textwidth]{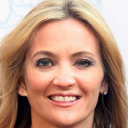} &
        \includegraphics[width=.145\textwidth]{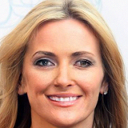}\\
        
        \includegraphics[width=.145\textwidth]{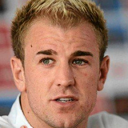} &   
        \includegraphics[width=.145\textwidth]{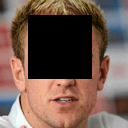} &
        \includegraphics[width=.145\textwidth]{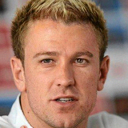} & 
        \includegraphics[width=.145\textwidth]{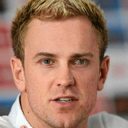} &
        \includegraphics[width=.145\textwidth]{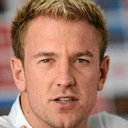} &
        \includegraphics[width=.145\textwidth]{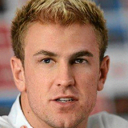}\\
        
        \includegraphics[width=.145\textwidth]{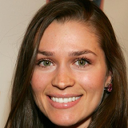} &   
        \includegraphics[width=.145\textwidth]{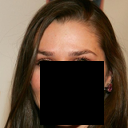} &
        \includegraphics[width=.145\textwidth]{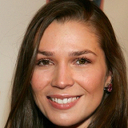} & 
        \includegraphics[width=.145\textwidth]{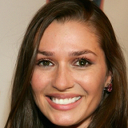} &
        \includegraphics[width=.145\textwidth]{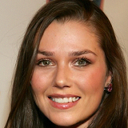} &
        \includegraphics[width=.145\textwidth]{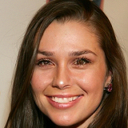}\\
        
        \includegraphics[width=.145\textwidth]{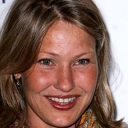} &   
        \includegraphics[width=.145\textwidth]{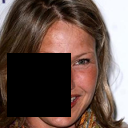} &
        \includegraphics[width=.145\textwidth]{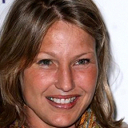} & 
        \includegraphics[width=.145\textwidth]{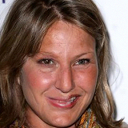} &
        \includegraphics[width=.145\textwidth]{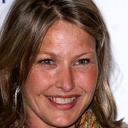} &
        \includegraphics[width=.145\textwidth]{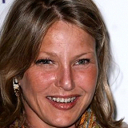}\\
        
        \includegraphics[width=.145\textwidth]{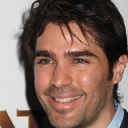} &   
        \includegraphics[width=.145\textwidth]{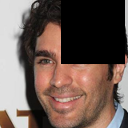} &
        \includegraphics[width=.145\textwidth]{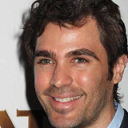} & 
        \includegraphics[width=.145\textwidth]{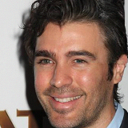} &
        \includegraphics[width=.145\textwidth]{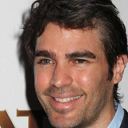} &
        \includegraphics[width=.145\textwidth]{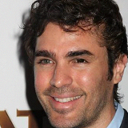}\\
        
        \includegraphics[width=.145\textwidth]{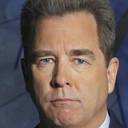} &   
        \includegraphics[width=.145\textwidth]{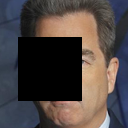} &
        \includegraphics[width=.145\textwidth]{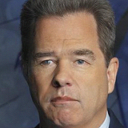} & 
        \includegraphics[width=.145\textwidth]{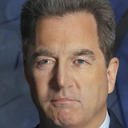} &
        \includegraphics[width=.145\textwidth]{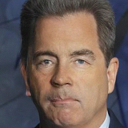} &
        \includegraphics[width=.145\textwidth]{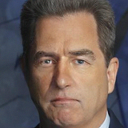}\\

    \end{tabular}
    \endgroup
    \end{center}
    \caption{Extended inpainting results.}
    \label{fig:additional_inpainting}
\end{figure*}

\begin{figure*}
    \begin{center}
    \begingroup
    \setlength{\tabcolsep}{2.5pt}
    \begin{tabular}{lcccccccc}
        \begin{tabular}{@{}l@{}}
            Original image $x$
            \\[25pt]
        \end{tabular}
         & \includegraphics[width=.08\textwidth]{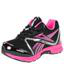} & \includegraphics[width=.08\textwidth]{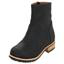} & \includegraphics[width=.08\textwidth]{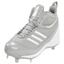} &
         \includegraphics[width=.08\textwidth]{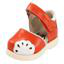} &
         \includegraphics[width=.08\textwidth]{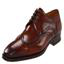} &
         \includegraphics[width=.08\textwidth]{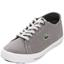} &
         \includegraphics[width=.08\textwidth]{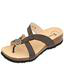} &
         \includegraphics[width=.08\textwidth]{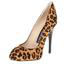} \\
        
        \begin{tabular}{@{}l@{}}
            Observation \\ $ y := Ax$
            \\[25pt]
        \end{tabular}
         & \includegraphics[width=.08\textwidth]{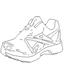} & \includegraphics[width=.08\textwidth]{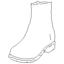} & \includegraphics[width=.08\textwidth]{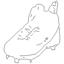} &
         \includegraphics[width=.08\textwidth]{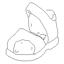} &
         \includegraphics[width=.08\textwidth]{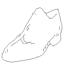} &
         \includegraphics[width=.08\textwidth]{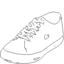} &
         \includegraphics[width=.08\textwidth]{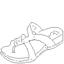} &
         \includegraphics[width=.08\textwidth]{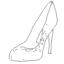} \\
        
         \begin{tabular}{@{}c@{}}
            CDE
            \\[25pt]
        \end{tabular} & 
            \includegraphics[width=.08\textwidth]{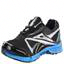} & \includegraphics[width=.08\textwidth]{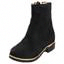} & \includegraphics[width=.08\textwidth]{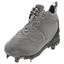} &
            \includegraphics[width=.08\textwidth]{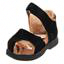} &
            \includegraphics[width=.08\textwidth]{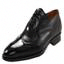} &
            \includegraphics[width=.08\textwidth]{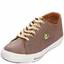} &
            \includegraphics[width=.08\textwidth]{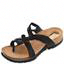} &
            \includegraphics[width=.08\textwidth]{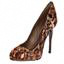} \\

        \begin{tabular}{@{}c@{}}
           CDiffE
            \\[25pt]
        \end{tabular} & 
            \includegraphics[width=.08\textwidth]{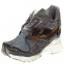} & \includegraphics[width=.08\textwidth]{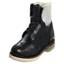} & \includegraphics[width=.08\textwidth]{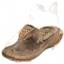} &
            \includegraphics[width=.08\textwidth]{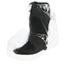} &
            \includegraphics[width=.08\textwidth]{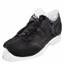} &
            \includegraphics[width=.08\textwidth]{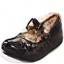} &
            \includegraphics[width=.08\textwidth]{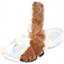} &
            \includegraphics[width=.08\textwidth]{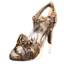} \\

        \begin{tabular}{@{}c@{}}
            CMDE
            \\[25pt]
        \end{tabular}  &  
            \includegraphics[width=.08\textwidth]{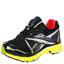} & \includegraphics[width=.08\textwidth]{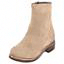} & \includegraphics[width=.08\textwidth]{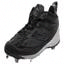} &
            \includegraphics[width=.08\textwidth]{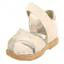} &
            \includegraphics[width=.08\textwidth]{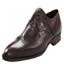} &
            \includegraphics[width=.08\textwidth]{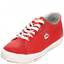} &
            \includegraphics[width=.08\textwidth]{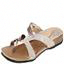} &
            \includegraphics[width=.08\textwidth]{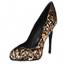} \\
            
        \begin{tabular}{@{}c@{}}
            VS-CMDE
            \\[25pt]
        \end{tabular}  &  
            \includegraphics[width=.08\textwidth]{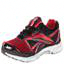} & \includegraphics[width=.08\textwidth]{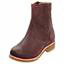} & \includegraphics[width=.08\textwidth]{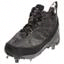} &
            \includegraphics[width=.08\textwidth]{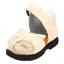} &
            \includegraphics[width=.08\textwidth]{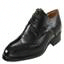} &
            \includegraphics[width=.08\textwidth]{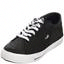} &
            \includegraphics[width=.08\textwidth]{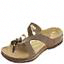} &
            \includegraphics[width=.08\textwidth]{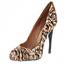} \\
    \end{tabular}
    \endgroup
    \end{center}
    \caption{Extended edge to shoe synthesis results.}
    \label{fig:additional_shoe}
\end{figure*}

We provide additional samples in Figures \ref{fig:additional_sr}, \ref{fig:additional_inpainting} and \ref{fig:additional_shoe}.

\section{Potential negative impact}
The potential of negative impact of this work is the same as that of any work that advances generative modeling. Generative modeling can be used for the creation of deep-fakes which can be used for malicious purposes such as disinformation and blackmailing. However, research on generative modeling can indirectly or directly contribute to the robustification of deep-fake detection algorithms. Moreover, generative models have proven very useful in academic research and in industry. The potential benefits of generative modeling outweigh the potential threats. Therefore, the research community should continue to conduct research on generative modeling.

\end{document}